\documentclass[letterpaper, 10 pt, conference]{ieeeconf}

\IEEEoverridecommandlockouts                              % This command is only needed if 
\usepackage{graphicx}
\usepackage{xcolor, colortbl}
\usepackage{array, multirow}
\usepackage{pgfplots}
\usepackage{pgfplotstable}
\usepackage{adjustbox}
\usepackage{comment}
\pgfplotsset{compat=1.14}
\usepackage{booktabs}
\usepackage{csquotes}
\usepackage{multirow}
\usepackage{siunitx}
\usepackage{textcomp}
\usepackage{float}
\usepackage{url}

\begin{document}
\title{\LARGE \bf IDDA: a large-scale multi-domain dataset for autonomous driving}

\author{Emanuele Alberti$^{*,1}$, 
        Antonio Tavera$^{*,1}$,
        Carlo Masone$^{2}$ 
        and Barbara Caputo$^{1}$% <-this % stops a space
%\thanks{Manuscript received: February 24, 2020; Revised: May 22, 2020; Accepted: June 22, 2020.} %Use only for final RAL version
%\thanks{This paper was recommended for publication by Editor Tamim Asfour upon evaluation of the Associate Editor and Reviewers' comments.}
%\thanks{This work was supported by Italdesign Giugiaro S.p.A.} %Use only for final RAL version
\thanks{*The authors equally contributed.}
\thanks{$^{1}$Emanuele Alberti, Antonio Tavera and Barbara Caputo are with Politecnico di Torino, Department of Control and Computer Engineering (DAUIN), Turin, Italy
        {\tt\footnotesize \{emanuele.alberti, antonio.tavera, barbara.caputo\}@polito.it}}%
\thanks{$^{2}$Carlo Masone is with Italdesign Giugiaro S.p.A., Turin, Italy {\tt\footnotesize carlo.masone@italdesign.it}}%
%\thanks{Digital Object Identifier (DOI): see top of this page.}
}

\maketitle

\begin{abstract}
Semantic segmentation is key in autonomous driving. Using deep visual learning architectures is not trivial in this context, because of the challenges in creating suitable large scale annotated datasets. 
This issue has been traditionally circumvented through the use of synthetic datasets, that have become a popular resource in this field. 
They have been released with the need to develop semantic segmentation algorithms able to close the visual domain shift between the training and test data. Although exacerbated by the use of artificial data, the problem is extremely relevant in this field even when training on real data. Indeed, weather conditions, viewpoint changes and variations in the city appearances can vary considerably from car to car, and even at test time for a single, specific vehicle. How to deal with domain adaptation in semantic segmentation, and how to leverage effectively several different data distributions (source domains) are important research questions in this field. To support work in this direction, this paper contributes a new large scale, synthetic dataset for semantic segmentation with more than 100 different source visual domains. The dataset has been created to explicitly address the challenges of domain shift between training and test data in various weather and view point conditions, in seven different city types. Extensive benchmark experiments assess the dataset, showcasing open challenges for the current state of the art. The dataset will be available at: \url{https://idda-dataset.github.io/home/}.
\end{abstract}

% \fi
%\IEEEpeerreviewmaketitle

\section{Introduction}
With the latest advancements 
in Deep Learning, we are starting to see a glimpse of what the future of the automotive industry might look like: self-driving cars that increase travel safety, reducing, if not nullifying, accidents. To achieve this ambitious goal, cars need to be aware of the environment that surrounds them in order to take the most appropriate action in each different situation. 

\begin{figure}[htbp]
    \centering
    \includegraphics[width=1.0\columnwidth]{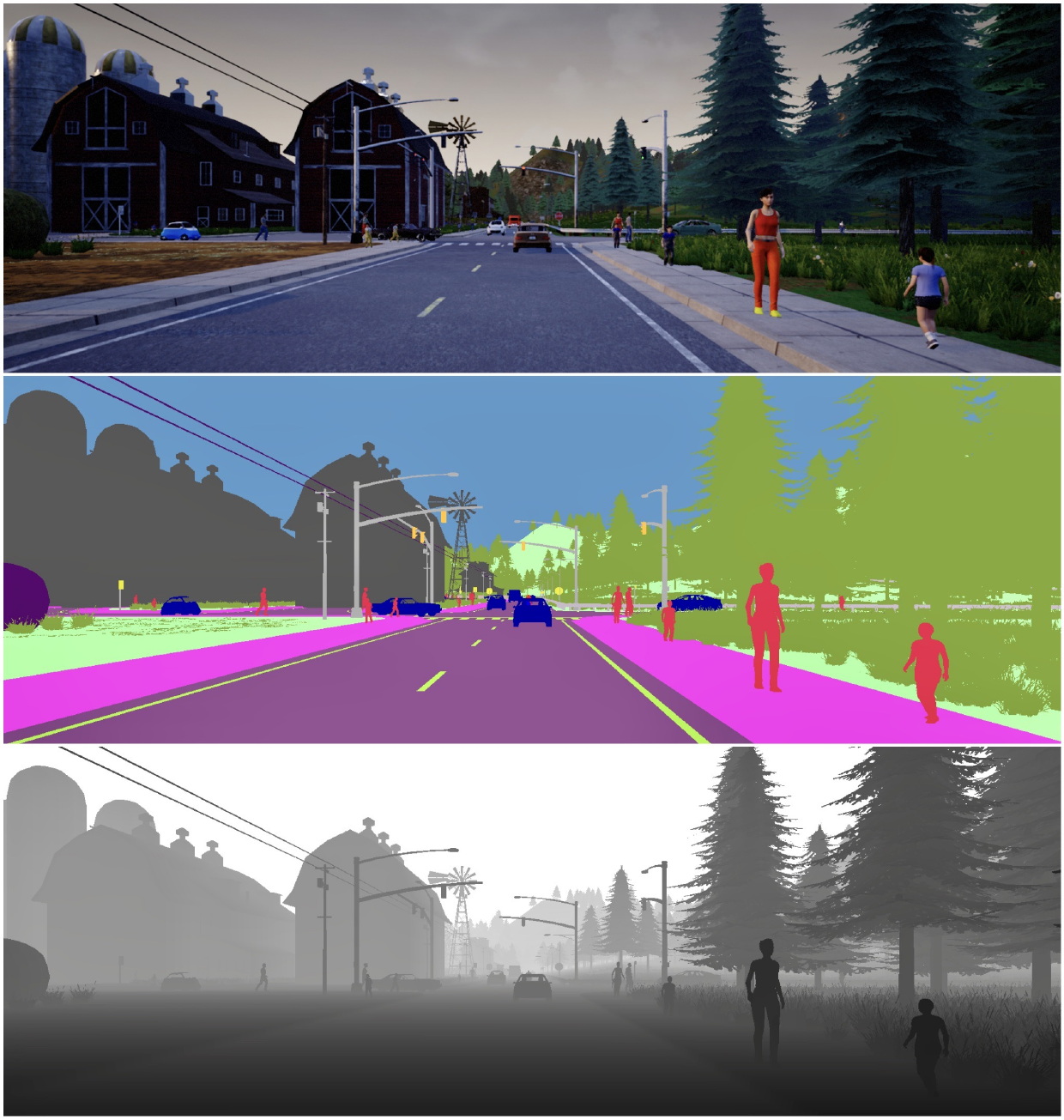}
    \caption{The IDDA dataset. An example with an RGB image and its corresponding semantic and depth maps.}
    \label{fig:IntroImage}
\end{figure}

\begin{table*}[t]
%\bigskip
\caption{Summary of the most popular dataset for Semantic Segmentation
}

\begin{adjustbox}{width=1.0\textwidth}
\begin{tabular}{@{}llllllllllllll@{}}
\toprule
\multicolumn{6}{l}{} &
  \multicolumn{3}{c}{\textbf{Semantic Segmentation}} &
  \multicolumn{5}{c}{\textbf{Data Variety}} \\ \midrule
\textbf{Dataset} &
  \textbf{Year} &
  \textbf{Size} &
  \textbf{Depth} &
  \textbf{\begin{tabular}[c]{@{}l@{}}Resolution\\ (pixels)\end{tabular}} &
  \textbf{FoV} &
  \textbf{\#Classes} &
  \textbf{\begin{tabular}[c]{@{}l@{}}Annotation\\ Time (min)\end{tabular}} &
  \textbf{\begin{tabular}[c]{@{}l@{}}\#Annotated\\ Pixels (10\textsuperscript{9})\end{tabular}} &
  \textbf{\begin{tabular}[c]{@{}l@{}}\#Weather\\ Conditions\end{tabular}} &
  \textbf{\#Envs} &
  \textbf{\#Viewpoints} &
  \textbf{\begin{tabular}[c]{@{}l@{}}\#Selectable \\Domains \end{tabular}} &
  \textbf{\begin{tabular}[c]{@{}l@{}}\#images\\ (avg*scene)\end{tabular}} \\ \midrule
\multicolumn{14}{c}{\textbf{Real-World Dataset}} \\ \midrule
CamVid &
  2008 &
  701 &
  No &
  920$\times$720 &
  - &
  32 &
  60 &
  0.62 &
  1 &
  1 &
  1 &
  1 &
  - \\ \midrule
KITTI &
  2012 &
  400 &
  \textbf{Yes} &
  1392$\times$512 &
  - &
  33 &
  - &
  0.07 &
  1 &
  1 &
  1 &
  1 &
  - \\ \midrule
Cityscapes &
  2016 &
  \begin{tabular}[c]{@{}l@{}}5k fine\\ 20k coarse\end{tabular} &
  No &
  2048$\times$1024 &
  90° &
  33 &
  \begin{tabular}[c]{@{}l@{}}90\\ 7\end{tabular} &
  \begin{tabular}[c]{@{}l@{}}9.43\\ 26.0\end{tabular} &
  - &
  \textbf{50} &
  1 &
  50 &
  160 \\ \midrule
Mapillary Vistas &
  2017 &
  25k &
  No &
  $\geq$ 1920$\times$1080 &
  - &
  \textbf{66} &
  94 &
  - &
  - &
  - &
  - &
  1 &
  - \\ \midrule
BDD100K &
  2018 &
  10k &
  No &
  1280$\times$720 &
  - &
  40 &
  - &
  - &
  6 &
  4 &
  1 &
  1 &
  - \\ \midrule
ApolloScape &
  2018 &
  144k &
  \textbf{Yes} &
  \textbf{3384$\times$2170} &
  - &
  25 &
  - &
  - &
  - &
  1 &
  1 &
  3 &
  \textbf{29k} \\ \midrule
A2D2 &
  2019 &
  41k &
  No &
  1920$\times$1280 &
  \textbf{120°} &
  38 &
  - &
  - &
  - &
  3 &
  6 (different horizontal position) &
  23 &
  1.7k \\ \midrule
\multicolumn{14}{c}{\textbf{Synthetic Dataset}} \\ \midrule
Virtual KITTI &
  2016 &
  21260 &
  Yes &
  $1242\times$375 &
  29° &
  14 &
  - &
  - &
  5 &
  5 &
  4 (different horizontal rotation) &
  50 &
  426 \\ \midrule
\begin{tabular}[c]{@{}l@{}}Synthia-Rand\\ Synthia-Seqs\end{tabular} &
  2016 &
  \begin{tabular}[c]{@{}l@{}}13,400\\ 200k\end{tabular} &
  \textbf{Yes} &
  960$\times$720 &
  100° &
  13 &
  \textbf{Instant} &
  147.5 &
  \begin{tabular}[c]{@{}l@{}}-\\ \textbf{10}\end{tabular} &
  4 &
  \textbf{8} (different horizontal position) &
  \begin{tabular}[c]{@{}l@{}}1\\ 51\end{tabular} &
  \begin{tabular}[c]{@{}l@{}}-\\ 8k\end{tabular} \\ \midrule
GTA V &
  2016 &
  25k &
  No &
  1914$\times$1052 &
  - &
  19 &
  7 &
  50.15 &
  - &
  1 &
  - &
  1 &
  - \\ \midrule
\textbf{IDDA} &
  \textbf{2020} &
  \textbf{1M} &
  \textbf{Yes} &
  1920$\times$1080 &
  90° &
  24 &
  \textbf{Instant} &
  \textbf{2087.70} &
  3 &
  7 &
  5 (different camera heights) &
  \textbf{105} &
  16k \\ \bottomrule
\end{tabular}
\end{adjustbox}
\vspace{-2mm}
\label{table:datasetComparison}
\end{table*}
Even though object detection/recognition based approaches \cite{od_survey} are very precise and reliable in some cases, they are not enough to accomplish such objective. 
A more profound comprehension of the scene is necessary if we want fine-graned decisions capabilities, e.g. deciding to go against a fence instead of a wall after a maneuver done to avoid a vehicle or a person.

Semantic Segmentation (SemSeg) \cite{semseg_survey} is a technology that, by classifying each individual pixel of the scene instead of just recognizing the main actors (such as vehicles and pedestrians), can enable driving systems to reach a better understanding of the whole view. Given the broad variety of driving conditions encountered in the real world, it is of paramount importance for these SemSeg algorithms to be able to generalize well and also cope with the inevitable domain shifts. On one side, this implies developing more effective domain adaptation (DA) techniques \cite{da_survey}  that are able to cope with such a diversity of unpredictable scenarios. On the other, this requires datasets with a large amount of labeled data from a diverse set of conditions to support the training and evaluation of such techniques.  

However, obtaining real labeled data in large quantities is far from trivial.
Firstly, it is both arduous and costly to deploy multiple vehicles to collect images from a multitude of weather, lighting and environmental conditions.

Secondly, the task of manually classifying each image is excessively time-consuming, with a duration that can range from 60 to 90 minutes per image, as it was for the CamVid \cite{camvid-1, camvid-2} and Cityscapes \cite{cityscapes_paper} datasets respectively. 
Lastly, the accuracy of the manually produced labels might be inconsistent throughout the dataset.

All these reasons, together with the level of fidelity reached by 3D graphical engines, have fostered the creation and adoption of synthetic datasets for SemSeg 
\cite{gta5, synthia}. 
Furthermore, automatically producing the labels directly from the objects in the 3D engine allows to have perfect labeling and to easily add new classes. 
The downside of this approach is that models trained solely on virtual datasets have the tendency to perform very poorly in real case scenarios, suffering from the domain shift, even though ways to tackle these issues are being developed in the form of domain adaptation and generalization.

In our work we propose ``IDDA'' (ItalDesign DAtaset), a 
large synthetic dataset counting over one million labeled images as the sum of
more than a hundred different scenarios over three axes of variability: 5 viewpoints, 7 towns and 3 weather conditions. 
The variety it offers allows for a deeper analysis and benchmarking of the performances of the current and future state-of-the-art SemSeg architectures, with a strong focus on DA  
tasks. 
For these reasons we believe that our dataset can bring a valuable contribution to the research community. The dataset, the experimental setups and all the algorithms used in this paper will be made publicly available through the dedicated webpage. The webpage will be periodically updated with new results and benchmark settings, with the explicit intention to make IDDA the reference resource for studying domain adaptive SemSeg in the automotive scenario.

To summarize, the main contributions of this paper are:
\begin{itemize}
    \item the creation of the largest synthetic dataset for semantic segmentation currently available, featuring more than 1M images, more than 100 different combinations of scenarios, and fine pixel-wise semantic annotations and depth maps.
    The scenarios are well-divided using the three variability factors: weather condition, location and camera height.
    \item the evaluation of the performances of the current state-of-the-art SemSeg models with their DA variants, assessing how useful the dataset proves to be for 
    benchmarking purposes, especially for a single-source DA task. We demonstrate how our dataset could potentially be employed to evaluate other tasks, such as multi-source DA or domain generalization.
\end{itemize}

\section{Related Work}
\begin{figure*}[ht]
    %\bigskip
    \centering
    \includegraphics[width=1.0\textwidth]{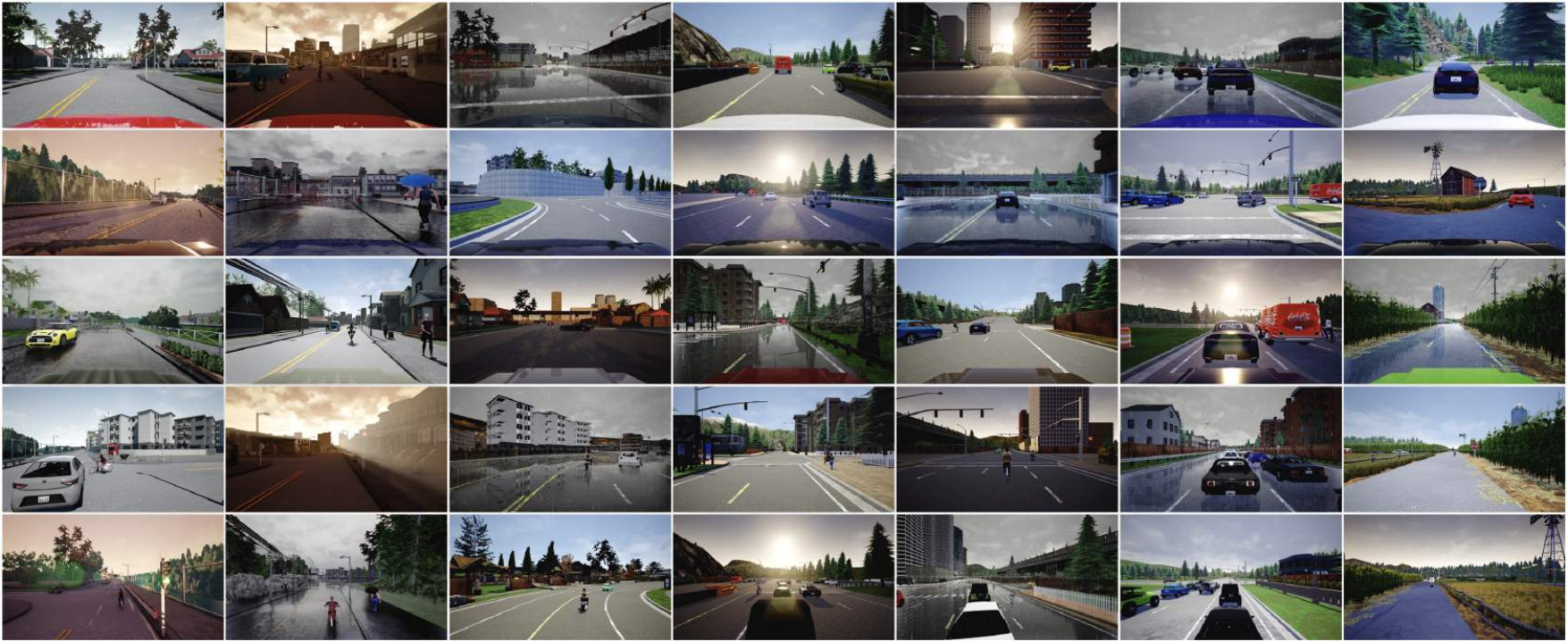}
    \vspace{-5mm}
    \caption{Samples for any instance of variety provided by the IDDA dataset. On the row the 5 viewpoints (Audi, Mustang, Jeep, Volkswagen T2 and Bus), on the column the 7 environments (from Town1 to Town7). Images iterate over the 3 weather conditions (Clear Noon, Clear Sunset and Hard Rain Noon).}
    \label{fig:scenariosExample}
\end{figure*}

The rapidly growing interest in the application of SemSeg to autonomous driving has led to the release of several datasets targeting this application (see Tab. \ref{table:datasetComparison}).
Early datasets, such as CamVid \cite{camvid-1,camvid-2} and KITTI \cite{kitti_paper}, while containing more than 30 classes of labeled objects, consisted of less than 1k semantically annotated images in low resolution and with little variability.
The release of Cityscapes \cite{cityscapes_paper}, with 5k finely annotated images and 20k coarsely annotated ones, led to the first benchmark to test SemSeg for autonomous driving.

The success of Cityscapes was later followed by the release of larger datasets from academic research (BDD100K \cite{bdd100k_paper}), image providers (Mapillary Vistas \cite{mapillary_paper}) and automotive industry (Apolloscape \cite{apolloscape_paper}, A2D2 \cite{audi_dataset}).
Despite the availability of multiple datasets, none of these has yet provided a good benchmark to evaluate how well a  
SemSeg network performs when tested on a different domain. Some datasets, such as CamVid, KITTI or Apolloscape, simply lack variability since they contain images taken from a single city or point of view. 
Others, such as Mapillary Vistas and BDD100K, that offer scene diversity but lack a way to easily pick scenarios from different domains, or Virtual KITTI \cite{virtual_kitti}, that provides few images per scenario, make it hard to use them to evaluate DA approaches.

The problem of collecting and labeling large quantities of images with a rich diversity of conditions has led to the creation of datasets based on 3D games engines such as SYNTHIA \cite{synthia} and GTA V \cite{gta5}. Using data from game engines also allows to get finely annotated images without the cost of manual labeling. Unfortunately, even these two datasets have limitations for what concerns their use to evaluate DA. GTA V does not currently offer the possibility of picking scenes from different domains whereas SYNTHIA-Seqs only contains low resolution images and few labels.
 
In comparison to these prior datasets IDDA is designed to provide a benchmark to test not only the generalization capabilities of SemSeg architectures, but also to assess how well they adapt to a domain shift. Our large-scale dataset consists of more than 1 FHD million images and it offers multiple domains easily and separately selectable. Together with each RGB image the dataset contains also its respective depth map and its high-quality semantic annotation for a total of 24 semantic classes, as shown in Fig. \ref{fig:IntroImage}.

\section{Data Creation} 
\subsection{The virtual simulator}
The simulator used for the generation of the dataset is CARLA \cite{carla_simulator} (version 0.8.4 and 0.9.6), an open-source project developed 
to support prototyping, training, and validation of autonomous driving systems. 
The motivation behind the choice of this simulator is the high degree of customization that it offers: the developer can set the number of pedestrians and vehicles, the environment conditions, the map and the speed of the simulation.
Moreover, CARLA uses Unreal Engine 4 which is the current state-of-the-art in computer graphics. 
From a practical perspective, CARLA is based on a client-server architecture, where the client controls a chosen individual agent (\emph{player}) while the server simulates the world and the remaining agents. This split allowed us to focus on implementing a custom made data-collection client without rewriting the server.

\subsection{Data-Collection Client}
Our client can start new simulations (\emph{episodes}), defining each time the parameters and the meta-parameters. 
The number of frames captured by the player in each episode is limited by the client depending on the size of the town: the smaller the town the fewer the images (i.e. the shorter the episodes).
Furthermore, to create different traffic scenarios, each episode is initialized with a random number of vehicles and pedestrians in the range of [20, 150] and [0, 100], respectively.
Lastly, players are spawned in new locations and in each episode the distributions of the vehicle models and colors keep changing.
These choices were made to limit the occurrence of deadlocks and, thus, the times in which the ego vehicle is stationary for any reason. Overall, these factors ensure that the collected data is rich and diverse. 

The client also specifies the sensors equipped on the player vehicle. Out of all the sensors available in CARLA, for the creation of the dataset we used an RGB camera, a semantic segmentation sensor and a depth sensor, all with a field-of-view (FoV) of 90 degrees. The semantic segmentation sensor produces instantly pixel-wise labeled images directly from the blueprints of the objects in the Unreal Engine.
The depth sensor provides images that codify depth in the 3 channels of the RGB color space, from the least to the most significant bytes: R \textgreater \space G \textgreater \space B. 

\begin{comment}
The actual distance in meters is calculated using the following formula:
\[ 
    \text{distance} = 1000\times\frac{R + G \times 256 + B \times 256^2}{256^3 - 1}
\]
\end{comment} 

The sensors are mounted coincidentally on the player's windshield, roughly at the height of the rear-view mirror.
Since we used 5 different player vehicle models to collect the data (two sport cars, a jeep, a minivan and a bus), the camera height ranges between 1.2 and 2.5 meters. Additionally, the portion of the image occupied by the player's hood varies depending on the model of the vehicle, ranging from 11.08\% to 13.99\% when the hood is visible (sedans and jeep) and equaling 0\% in the other cases.

All sensors are synchronized to capture a frame every 3 seconds, leading to episodes lasting from 3 to 4 minutes each (simulation time).

\begin{figure}[ht]
    %\bigskip
    \centering
    \includegraphics[width=1.0\columnwidth]{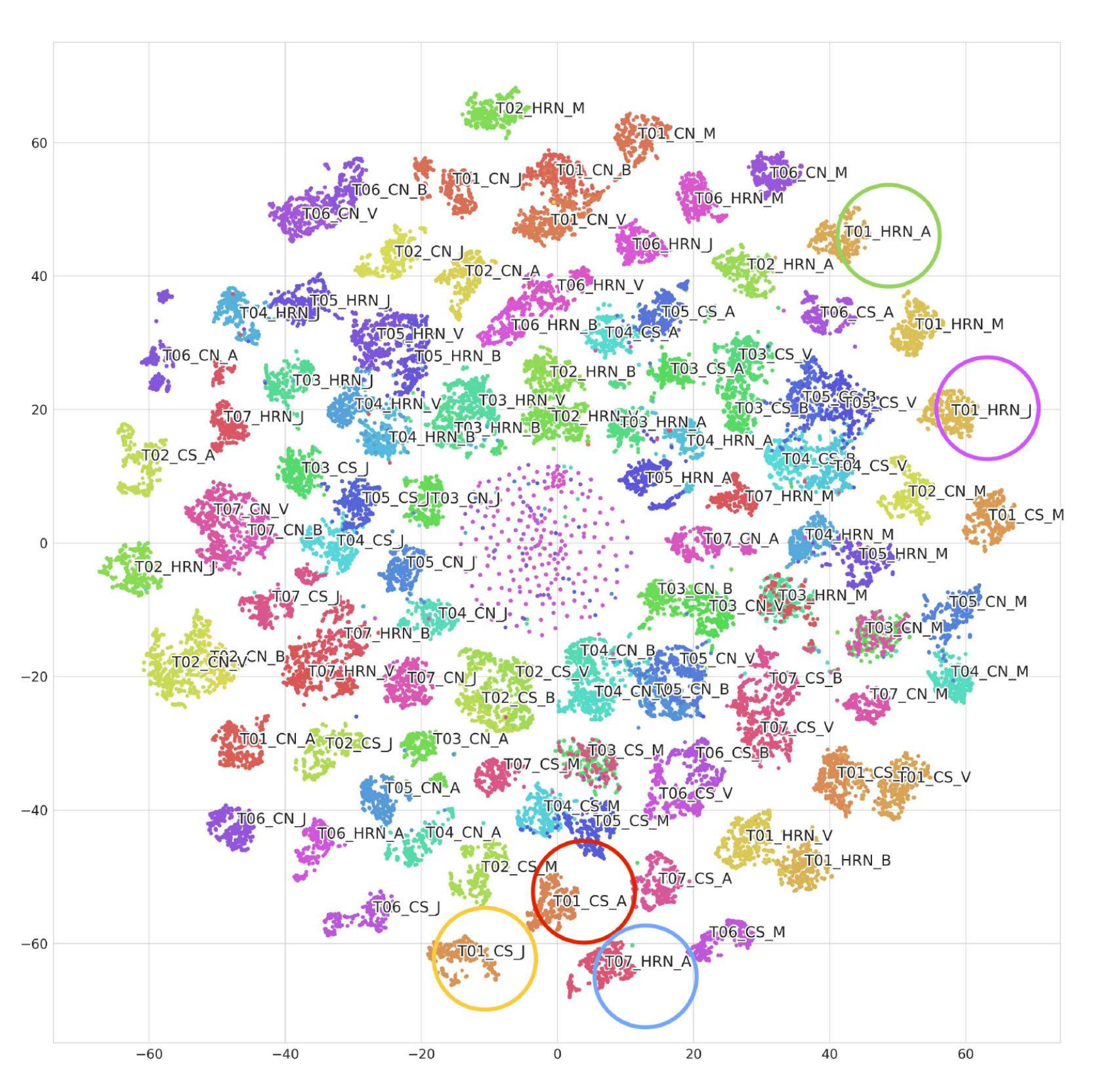}
    \vspace{-5mm}
    \caption{The tSNE representation of the 105 different IDDA's scenarios. Inside the circles are the domains tested in Sec. \ref{subsection:assessing_idda}.}
    \label{fig:105classes}
\end{figure}

At the moment of capture, six frames are simultaneously stored: one RGB, three depth (raw, grayscale, and log-grayscale), two semantic (raw and colored using the Cityscapes color palette).
For the RGB camera, post-processing effects such as bloom, lens flare and motion blur are applied in order to increase the realism of the images.

\section{The Dataset}
IDDA (ItalDesign DAtaset) consists of 1,006,800 frames taken from the virtual world simulator CARLA.
The creation of the dataset lasted about two weeks on two workstations, each equipped with a  single NVIDIA Quadro P5000 GPU with 16GB of memory. In terms of quantity of frames, IDDA is 2 orders of magnitude larger than GTAV \cite{gta5} and SYNTHIA \cite{synthia} and 5 order of magnitude larger than semantically annotated images in KITTI \cite{kitti_paper}.
Most importantly, IDDA features many scenarios spanning different cities, weather conditions and viewpoints, so as to support the development and evaluation of single or multi-source DA techniques applied to SemSeg.

\subsection{Data Diversity}
The 105 scenarios composing IDDA (examples in Fig. \ref{fig:scenariosExample})
are obtained by varying three aspects of the simulation.

\textbf{Towns.} The frames of the dataset are collected across seven different towns. Town1 (T01) and Town2 (T02) are characterized respectively by 2.9 km and 1.4 km of drivable roads with buildings, bridges, vegetation, terrain, traffic signs and various kinds of infrastructures. Town3 (T03), Town4 (T04), Town5 (T05) and Town6 (T06) are characterized by a complex urban scene with multi-lane roads, tunnels, roundabouts, freeways and connection ramps. Lastly, Town7 (T07) stands out from the rest because it depicts a bucolic countryside with narrow roads, fewer traffic lights and lots of non-signalized crossings.
We believe that this entirely different domain is one important novelty provided by our dataset with regards to the autonomous driving task.  All the seven cities are populated by vehicles and pedestrians.

\textbf{Weather Conditions.} 
We considered three weather settings that differ significantly from each other: Clear Noon (CN), characterized by bright daylight, Clear Sunset (CS), with the sun low above the horizon and pink/orange hues, and Hard Rain Noon (HRN), with a cloudy sky, intense rain and puddles that cause reflections on the floor.

\textbf{Viewpoints.} The third parameter that is varied to create the scenarios is the player vehicle. For each vehicle we positioned the sensor system approximately at the height of the rear-view mirror. We used five player vehicles that differ significantly in their height and shape, i.e, an Audi TT (A), a Ford Mustang (M), a Jeep Wrangler (J), a Volkswagen T2 (V) and a Bus (B). This choice guarantees not only that the resulting images have distinct perspectives, but also that the hood of the player vehicle, if visible\footnote{The hood is not visible in the case of the Bus and the Volkswagen T2.}, is dissimilar in both shape and color.
To the best of our knowledge, the inclusion of images not only from the perspective of cars but also jeeps, vans and buses is a a unique feature of IDDA and it adds a whole new dimension of variability.

We use tSNE \cite{tsne} to visually examine and evaluate the diversity of all the 105 available scenarios.
To do so, we train a ResNet101 \cite{resnet} from scratch, using 1000 samples from each scenario, with the sole task of classifying the domain of origin for each frame.
Then, for each scenario, we compute its mean feature vector using 500 samples randomly taken from its validation set. Finally, we apply PCA, take the first 50 principal components and project them into a more intelligible 2D embedding. Fig.  \ref{fig:105classes} presents a drawing of this embedding that intuitively shows the inherent domain shift that exists among the different scenarios. There is a strict correlation between the gap observed in the distribution of the domains in Fig. \ref{fig:105classes} and the results in terms of mIoU. 
Even if at a different scale, the similarity with Fig. \ref{fig:tSNE4Scenarios}, in which tSNE is computed only for handpicked sub-domains, is clearly discernible. In the experiments section we will demonstrate that this shift is strictly related to the results.

\begin{figure}[t]
    \definecolor{unlabeled}             {RGB}{  0,  0,  0}
\definecolor{building}              {RGB}{  70, 70, 70}
\definecolor{fence}                 {RGB}{  190, 153, 153}
\definecolor{other}                 {RGB}{  72, 0, 90}
\definecolor{pedestrian}            {RGB}{  220, 20, 60}
\definecolor{pole}                  {RGB}{153, 153, 153}
\definecolor{road line}             {RGB}{ 157, 234, 50}
\definecolor{road}                  {RGB}{128, 64, 128}
\definecolor{sidewalk}              {RGB}{244, 35, 232}
\definecolor{vegetation}            {RGB}{107, 142, 35}
\definecolor{vehicle}               {RGB}{0, 0, 142}
\definecolor{wall}                  {RGB}{102, 102, 156}
\definecolor{traffic sign}          {RGB}{220, 220, 0}
\definecolor{traffic light}         {RGB}{250, 170, 30}
\definecolor{guard rail}            {RGB}{180, 165, 180}
\definecolor{dynamic}               {RGB}{111, 74, 0}
\definecolor{bicycle}               {RGB}{119, 11, 32}
\definecolor{motorcycle}            {RGB}{0, 0, 230}
\definecolor{rider}                 {RGB}{255, 0, 0}
\definecolor{terrain}               {RGB}{152, 251, 152}
\definecolor{sky}                   {RGB}{70, 130, 180}
\definecolor{railway}               {RGB}{230, 150, 140}
\definecolor{ground}                {RGB}{81, 0, 81}
\definecolor{static}                {RGB}{0, 0, 0}
%\bigskip
\newcommand{\fnn}[1]{$\scriptstyle^{#1}$}
\begin{adjustbox}{width=1.0\columnwidth} %use \textwidth if you want all page size or \columnwidth for column
\begin{tikzpicture}

\tikzstyle{every node}=[font=\large]
\begin{axis}[
        ybar,
        %ymode=normal,
        ymode=log, %se log
        width=\textwidth,
        height=7cm, %7cmm
        ymin=2.5e9, %se log
        ymax=2.5e16, %se log
        ytick={10e9, 10e11, 10e13, 10e15},
        ylabel={number of pixels}, %se log
        xlabel= \empty,
        xtick = \empty,
        xmin = 0,
        xmax = 25,
        ymajorgrids=true,
        every node near coord/.append style={
                anchor=west,
                rotate=90,
                font=\large,
        },
        legend image code/.code={
        \draw [/tikz/.cd,bar width=2pt,yshift=-0.2em,bar shift=0pt]
        plot coordinates {(0cm,0.8em)};
        },
        ]
\addlegendimage{black, only marks, mark=asterisk}
\addlegendentry{Classes not considered during evaluation}

\addplot[bar shift=0pt,draw=unlabeled, fill opacity=0.8,fill=unlabeled!80!white, nodes near coords=unlabeled\textsuperscript{*}] plot coordinates{ ( 1,  228166868105) };
\addplot[bar shift=0pt,draw=building, fill opacity=0.8,fill=building!80!white , nodes near coords=building] plot coordinates{ ( 2,     23307916655951  ) };
\addplot[bar shift=0pt,draw=fence,       fill opacity=0.8,fill=fence!80!white        , nodes near coords=fence       ] plot coordinates{ ( 3,    1779701571219  ) };
\addplot[bar shift=0pt,draw=other,    fill opacity=0.8,fill=other!80!white     , nodes near coords=other\textsuperscript{*}    ] plot coordinates{ ( 4,     1320683518913    ) };
\addplot[bar shift=0pt,draw=pedestrian,      fill opacity=0.8,fill=pedestrian!80!white       , nodes near coords=pedestrian               ] plot coordinates{ ( 5,    263062977344 ) };
\addplot[bar shift=0pt,draw=pole,         fill opacity=0.8,fill=pole!80!white          , nodes near coords=pole                ] plot coordinates{ ( 6,    1704540720549    ) };
\addplot[bar shift=0pt,draw=road line,          fill opacity=0.8,fill=road line!80!white           , nodes near coords=road line\textsuperscript{*}                 ] plot coordinates{ ( 7,    2034711600277    ) };
\addplot[bar shift=0pt,draw=road,        fill opacity=0.8,fill=road!80!white         , nodes near coords=road        ] plot coordinates{ ( 8,     49463892689073   ) };
\addplot[bar shift=0pt,draw=sidewalk,        fill opacity=0.8,fill=sidewalk!80!white         , nodes near coords=sidewalk        ] plot coordinates{ ( 9,    13314207832951) };
\addplot[bar shift=0pt,draw=vegetation,    fill opacity=0.8,fill=vegetation!80!white     , nodes near coords=vegetation    ] plot coordinates{ ( 10,    30960901843803      ) };
\addplot[bar shift=0pt,draw=vehicle,    fill opacity=0.8,fill=vehicle!80!white     , nodes near coords=vehicle               ] plot coordinates{ ( 11,    15990470980019 ) };
\addplot[bar shift=0pt,draw=wall,       fill opacity=0.8,fill=wall!80!white        , nodes near coords=wall              ] plot coordinates{ ( 12,    1726015249312   ) };
\addplot[bar shift=0pt,draw=traffic sign,           fill opacity=0.8,fill=traffic sign!80!white            , nodes near coords=traffic sign           ] plot coordinates{ ( 13,    246957080772   ) };
\addplot[bar shift=0pt,draw=traffic light,       fill opacity=0.8,fill=traffic light!80!white        , nodes near coords=traffic light       ] plot coordinates{ ( 14,    185217810579   ) };
\addplot[bar shift=0pt,draw=guard rail,           fill opacity=0.8,fill=guard rail!80!white            , nodes near coords=guard rail\textsuperscript{*}           ] plot coordinates{ ( 15,    585180908787    ) };
\addplot[bar shift=0pt,draw=dynamic,         fill opacity=0.8,fill=dynamic!80!white          , nodes near coords=dynamic\textsuperscript{*}         ] plot coordinates{ ( 16,   88582431146    ) };
\addplot[bar shift=0pt,draw=bicycle,         fill opacity=0.8,fill=bicycle!80!white          , nodes near coords=bicycle         ] plot coordinates{ ( 17,    24158844858    ) };
\addplot[bar shift=0pt,draw=motorcycle,    fill opacity=0.8,fill=motorcycle!80!white     , nodes near coords=motorcycle    ] plot coordinates{ ( 18,    93951063337   ) };
\addplot[bar shift=0pt,draw=rider,       fill opacity=0.8,fill=rider!80!white        , nodes near coords=rider     ] plot coordinates{ ( 19,    131531488672     ) };
\addplot[bar shift=0pt,draw=terrain,       fill opacity=0.8,fill=terrain!80!white        , nodes near coords=terrain     ] plot coordinates{ ( 20,    10117187363386   ) };
\addplot[bar shift=0pt,draw=sky,           fill opacity=0.8,fill=sky!80!white            , nodes near coords=sky                  ] plot coordinates{ ( 21,    54695624758916  ) };
\addplot[bar shift=0pt,draw=railway,          fill opacity=0.8,fill=railway!80!white           , nodes near coords=railway\textsuperscript{*}                 ] plot coordinates{ ( 22,442912155733   ) };
\addplot[bar shift=0pt,draw=ground,  fill opacity=0.8,fill=ground!80!white   , nodes near coords=ground\textsuperscript{*}         ] plot coordinates{ ( 23,    5368632190   ) };
\addplot[bar shift=0pt,draw=static, fill opacity=0.8,fill=static!80!white  , nodes near coords=static\textsuperscript{*}       ] plot coordinates{ ( 24,    59054954097) };
\end{axis}

\end{tikzpicture}
\end{adjustbox}
    \vspace{-5mm}
    \caption{Number of high-quality annotated pixels (y-axis) per class (x-axis).}
    \label{fig:pixeldistr}
\end{figure}
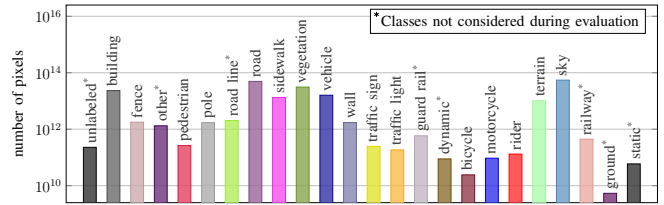

\subsection{Semantic Segmentation}
One of our goals in the creation of IDDA  was to build a competitive dataset in terms of the range of recognizable items within a scene. In particular, we wanted to increase the default number of semantic classes provided by the simulator and get it as close as possible to the ones in Cityscapes or in GTA V. In order to achieve this result we made changes in the 3D maps themselves and we modified and rebuilt the source code of the simulator so that each static and dynamic element would be identified and tagged the moment before being spawned inside the virtual world. This strategy allowed us to increase the number of tags provided by the simulator from the original 13 to a total number of 24 semantic classes. 
The distribution of classes in the IDDA dataset is analyzed in Fig. \ref{fig:pixeldistr}. It is clearly distinguishable that the predominant classes are building, road, vehicle, vegetation, terrain and sky. Other useful statistics are synthesized in the Tab. \ref{table:datasetComparison}.

\section{Experiments}
We demonstrate the main features and potential applications of IDDA with two experiments.
In the first one we want to verify that the scenarios available in IDDA are an effective tool to validate and benchmark how well SemSeg methods can adapt to domain shifts in driving applications. To do so, we selected several state-of-the-art networks, both with and without DA, and we looked at the performance degradation when the train and test sets are taken from different scenarios.
With the second experiment we use the scenarios available in IDDA to investigate how different data distributions in the synthetic source domain affect the performance of a network on a real target domain.
For this purpose we use the same networks from the first experiment but test them on Cityscapes, BDD100K, Mapillary Vistas and A2D2.

\textbf{Evaluated methods.} For the experiments we use eight state-of-the-art SemSeg architectures. Four of these networks do not implement DA, i.e. PSPNet \cite{pspnet}, that introduces a Pyramid Scene Parsing module, PSANet \cite{psanet}, that proposes a point-wise spatial attention network to gather information from all the positions in the feature maps and DeepLab V3+ \cite{deep-labv3+}, that implements the Atrous Spatial Pyramid Pooling module.
The fourth SemSeg architecture included in our experiments is DeepLab V2 \cite{deep-labv2} with a ResNet-101 \cite{resnet} as backbone, because this is the main building block for all the chosen DA methods.

The remaining four architectures are some of the best performing unsupervised DA models:
ADVENT \cite{advent}, DISE \cite{dise}, CLAN \cite{clan}, and DADA \cite{dada}. Each approach achieves its goal in a different way with respect to the others: both ADVENT and DADA use an entropy minimization technique with the help of an adversarial task, but the latter also takes advantage of depth information, DISE unravels images into domain-invariant structure and domain-specific texture representations, allowing for label transferring, and CLAN takes into account the local semantic consistency when pursuing the global alignment of the distributions, reducing the negative transfer side effect, that is the misalignment of features that were already aligned well prior to the mapping.

\textbf{Experimental setup.} \label{section:setup}
For each network we used the hyperparameters reported in its original paper, so as to obtain a fair evaluation of the performances.
For all the DA architectures 
the official implementation provided by the authors is used, whereas for the SemSeg-only part of the experiments re-implementations in PyTorch are used.
\begin{table}[htpb]
%\bigskip
\caption{Scenarios Distances}
\centering
\begin{tabular}{@{}lccc@{}}
\toprule
                  & \multicolumn{3}{c}{Case} \\ \midrule
Distance Function & Viewpoint Change   & Weather Change & City Change   \\ \midrule
Euclidean         & 2.7604 & 5.6555 & 6.4551 \\ \midrule
Cosine            & 0.2590 & 1.2633 & 1.0586 \\ \midrule
Bhattacharyya     & 0.0149 & 0.0337 & 0.0426 \\ \bottomrule
\end{tabular}
\vspace{-2mm}
\label{table:scenariosDistance}
\end{table}
\begin{figure}[htpb]
    \centering
    \includegraphics[width=1.0\columnwidth]{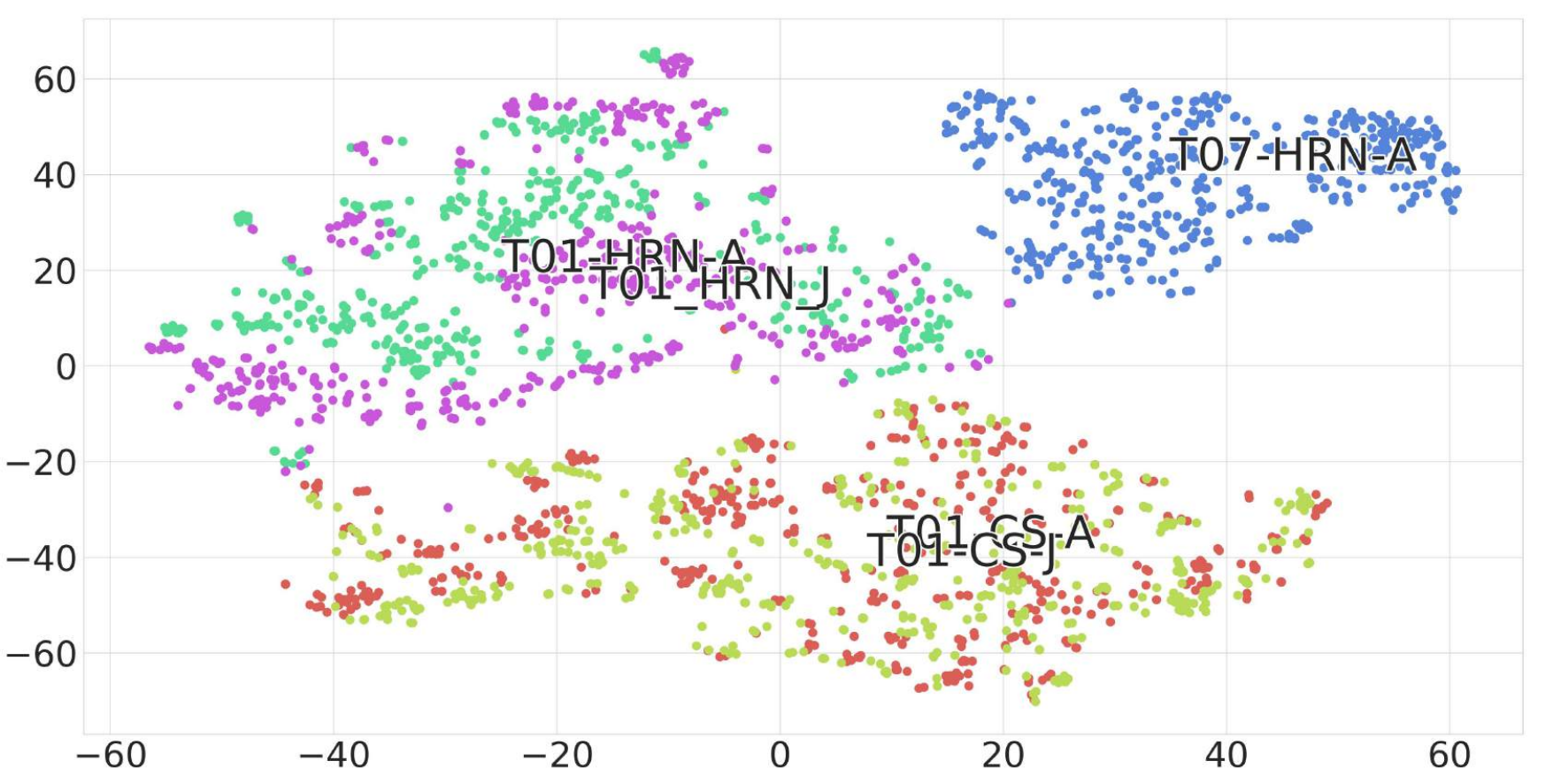}
    \vspace{-5mm}
    \caption{The tSNE representation of the 5 chosen scenarios to assess IDDA.}
    \label{fig:tSNE4Scenarios}
\end{figure}
\begin{figure*}[ht]
%\bigskip
\small
\begin{tabular}{ccccccc}
\centering
\multirow{4}{*}{\rotatebox{90}{Viewpoint Change\hspace{3.0mm}}} &
\raisebox{-0.5\height}{\includegraphics[width=29.5mm]{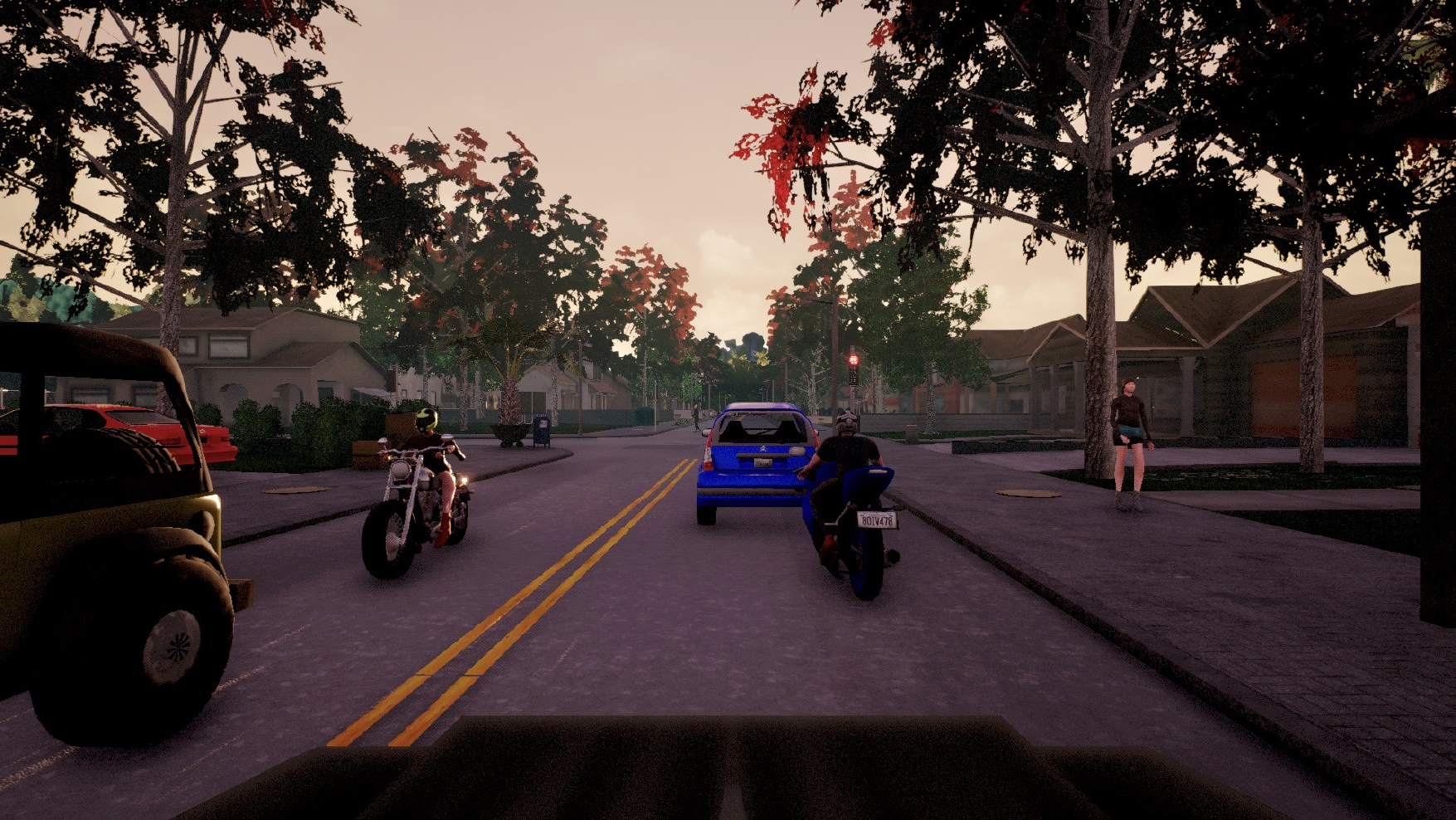}} &   &                \raisebox{-0.5\height}{\includegraphics[width=29.5mm]{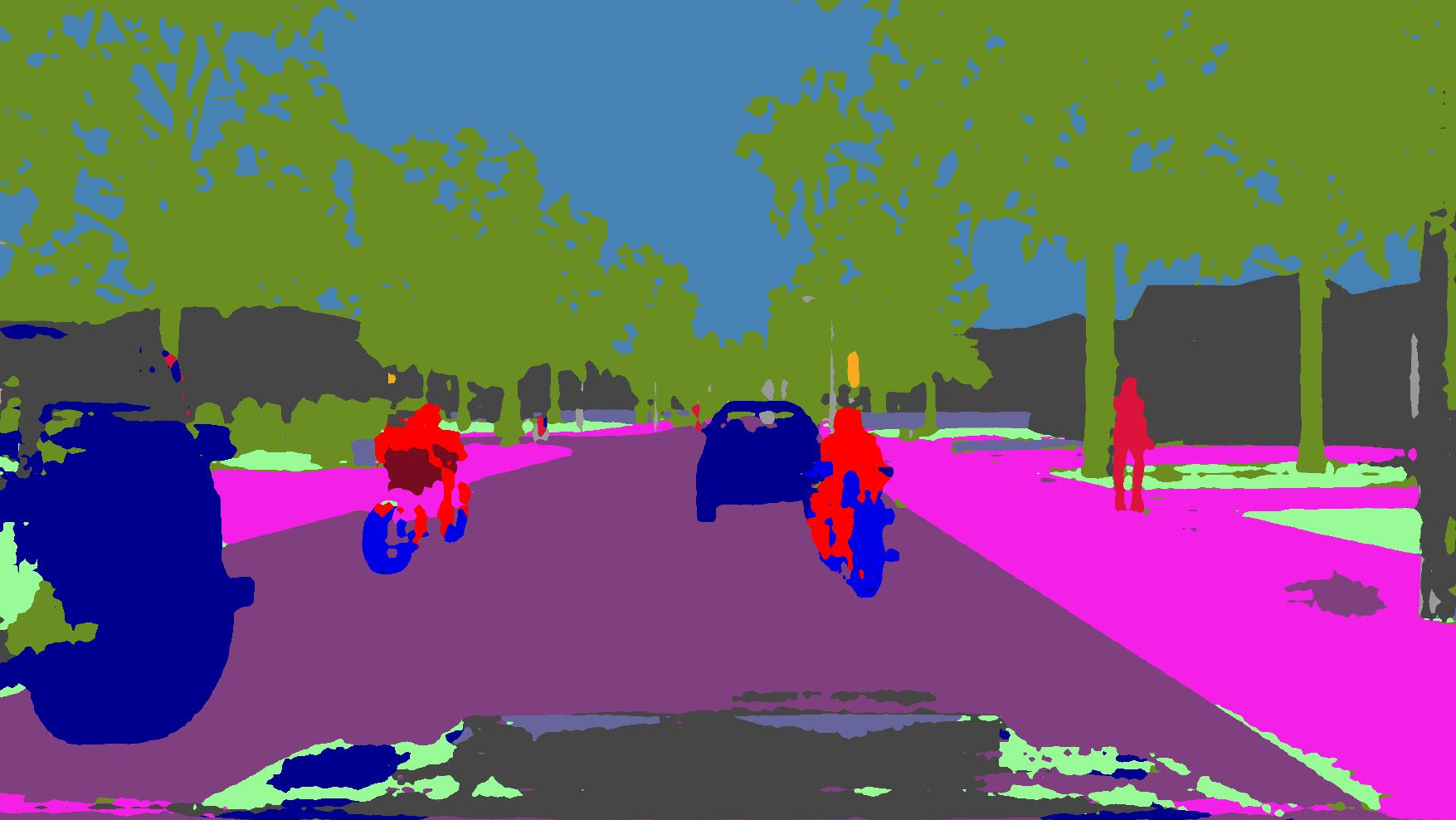}} & 
\raisebox{-0.5\height}{\includegraphics[width=29.5mm]{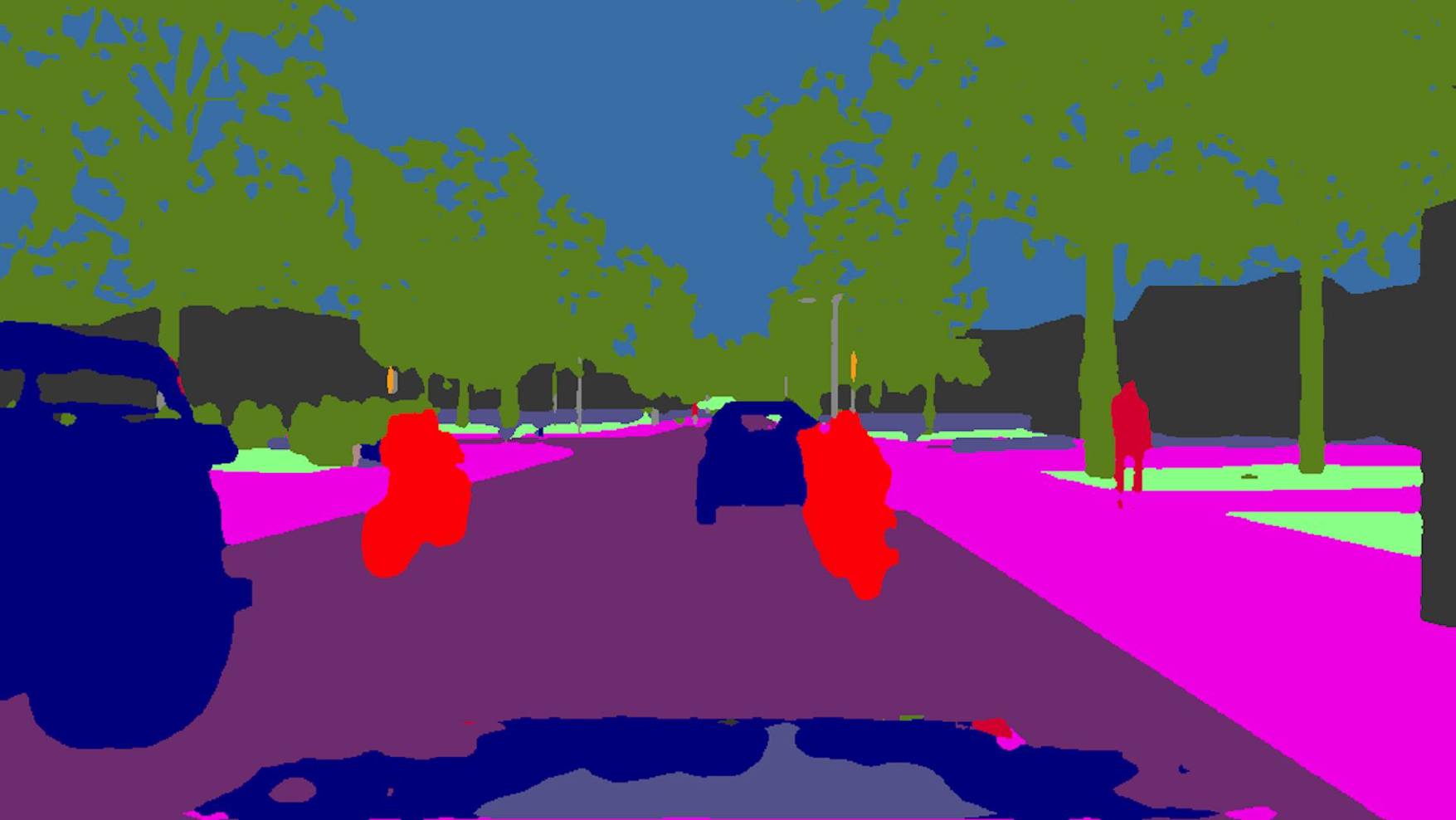}} & 
\raisebox{-0.5\height}{\includegraphics[width=29.5mm]{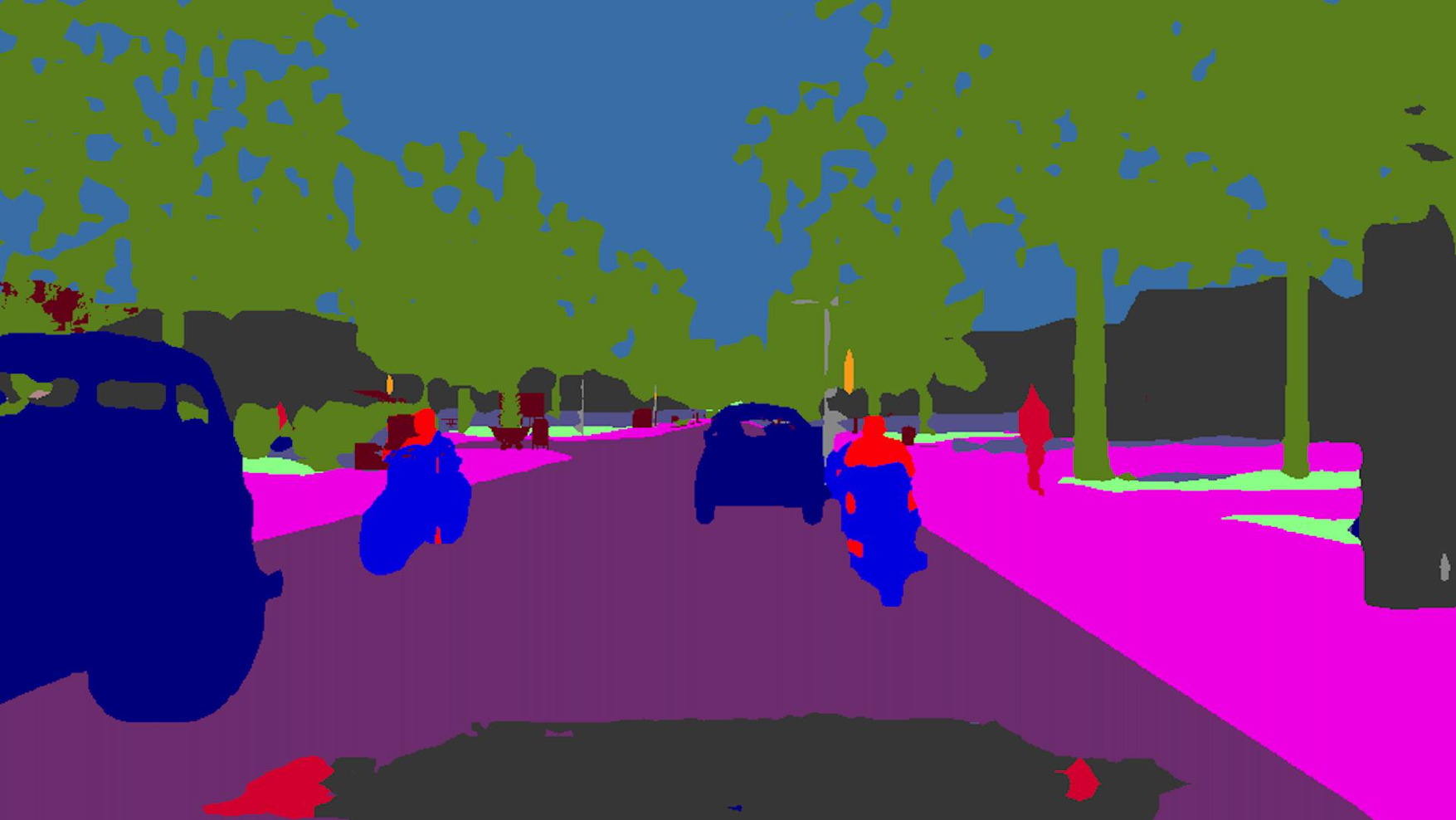}} &  
\raisebox{-0.5\height}{\includegraphics[width=29.5mm]{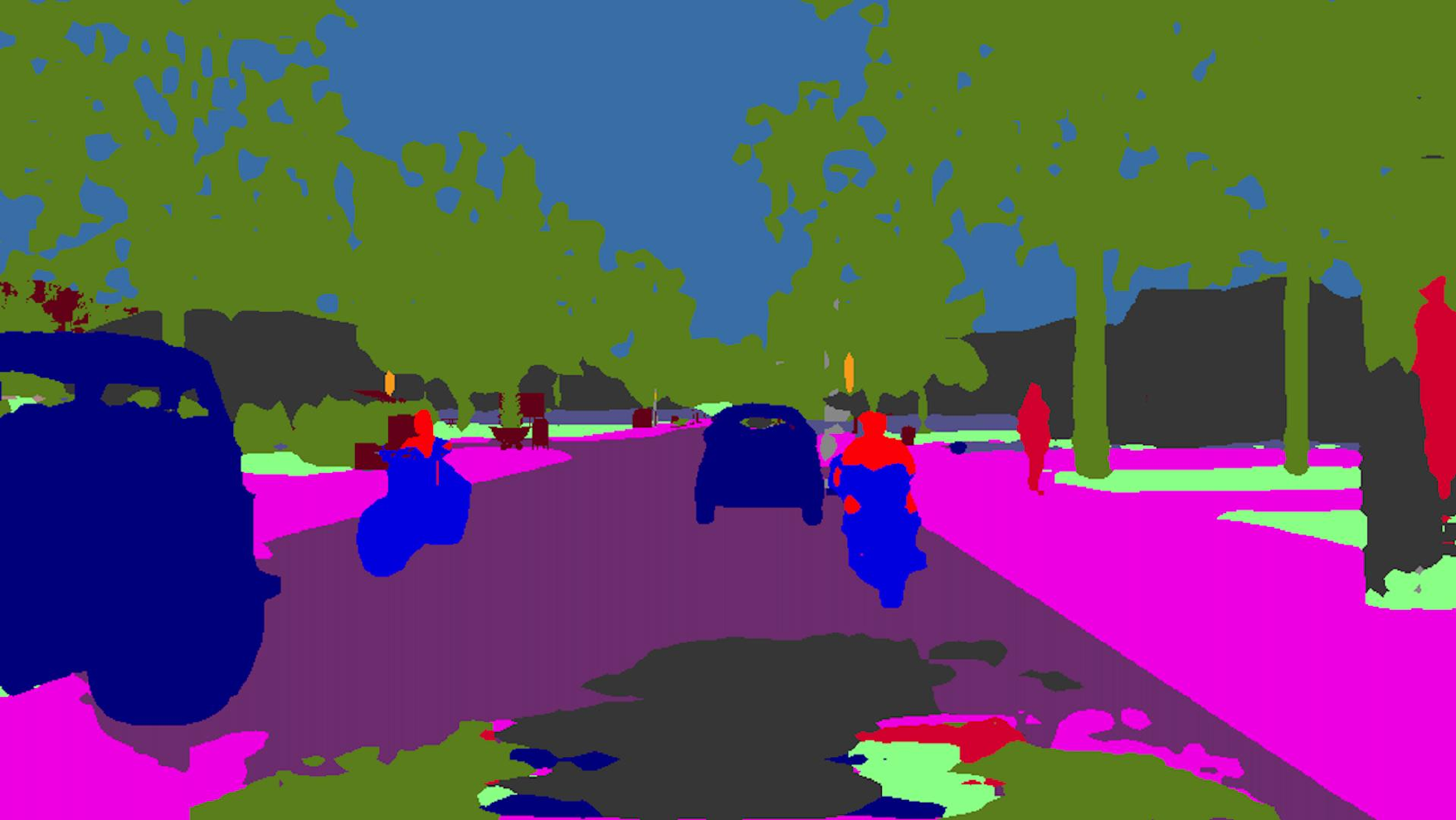}}\\&
a) RGB &   &                
c) DeepLab V2 & 
d) DeepLab V3+ & 
e) PSPNet &  
f) PSANet\\&
\raisebox{-0.5\height}{\includegraphics[width=29.5mm]{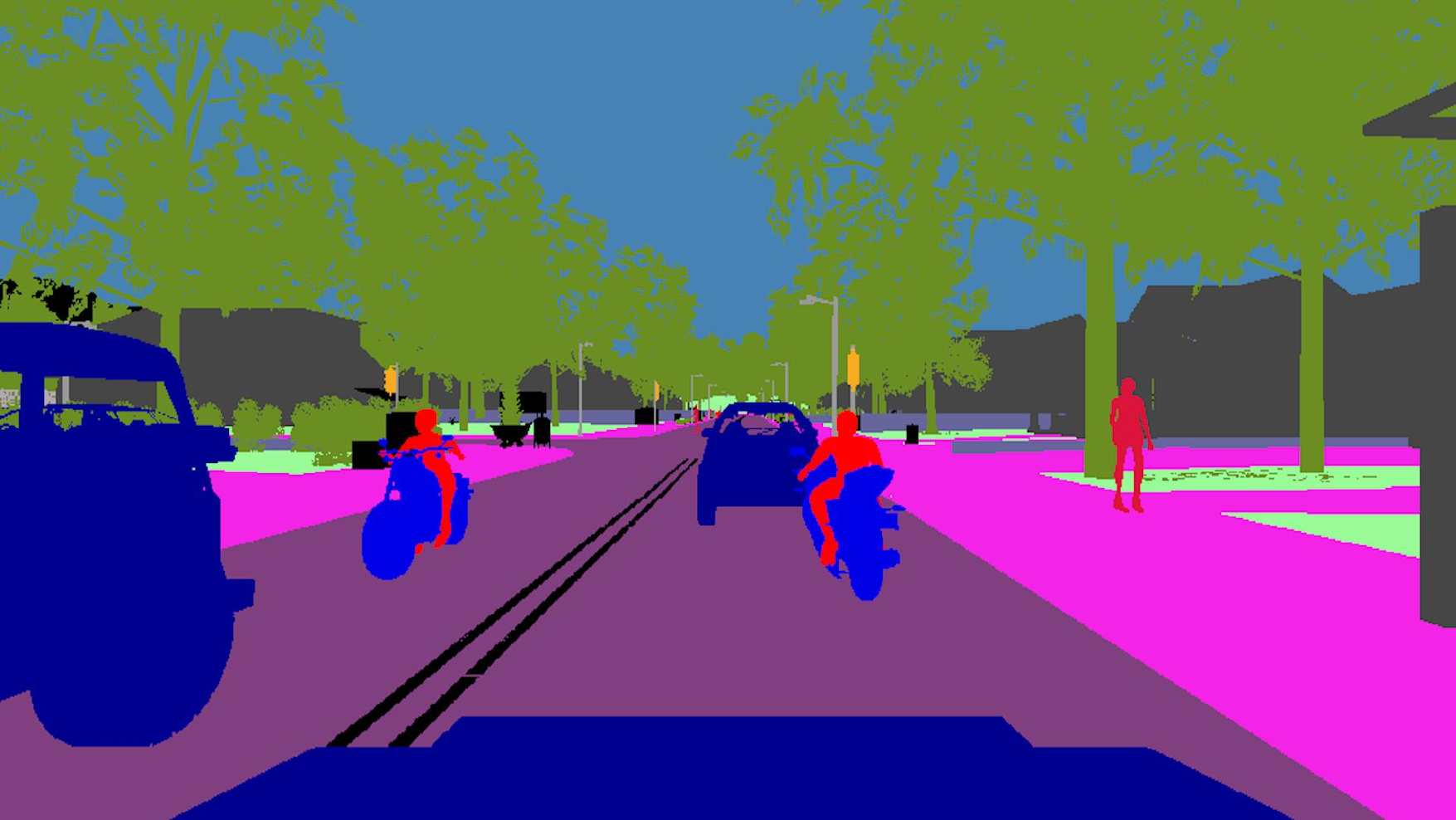}} & &
\raisebox{-0.5\height}{\includegraphics[width=29.5mm]{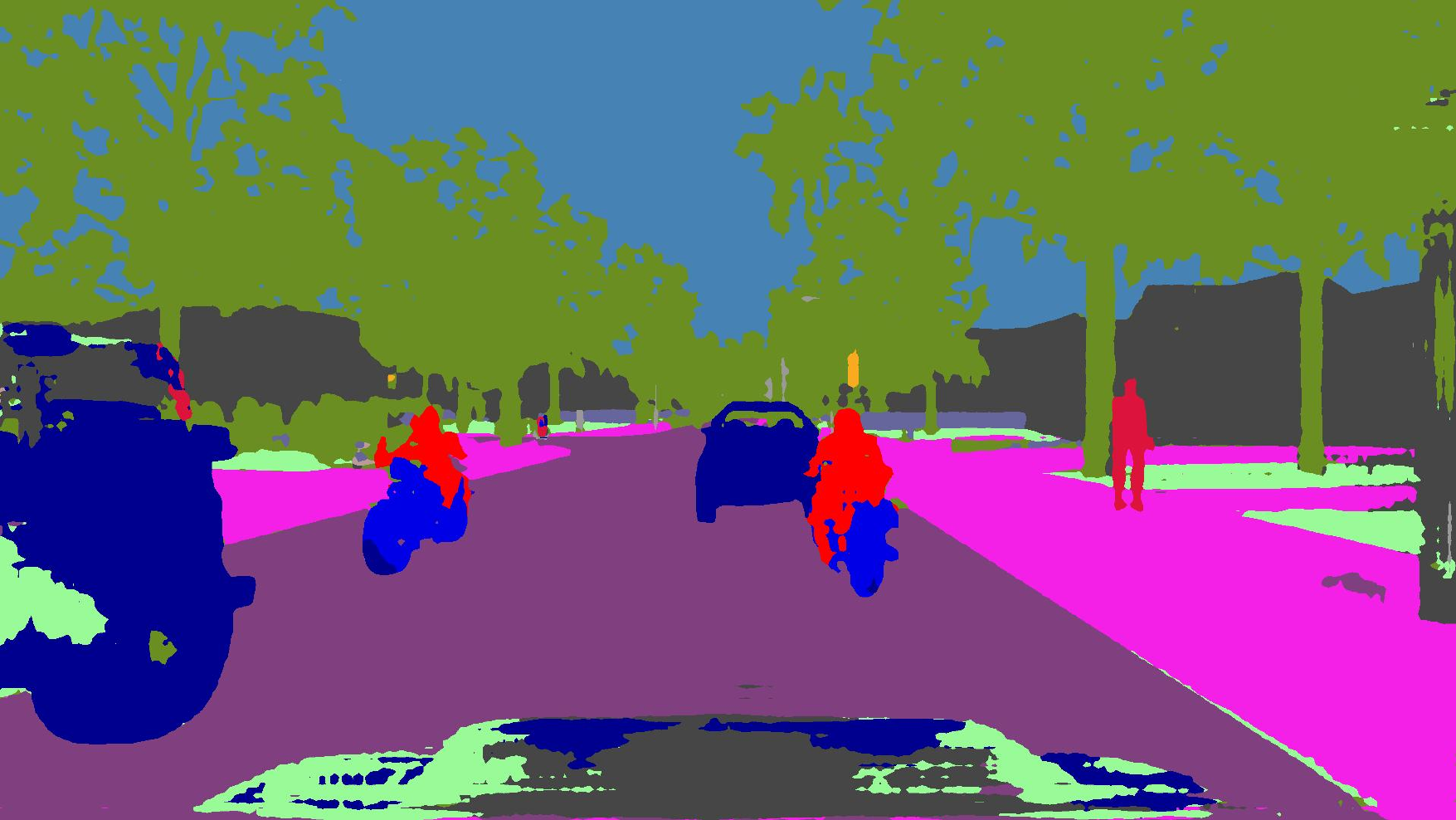}} &
\raisebox{-0.5\height}{\includegraphics[width=29.5mm]{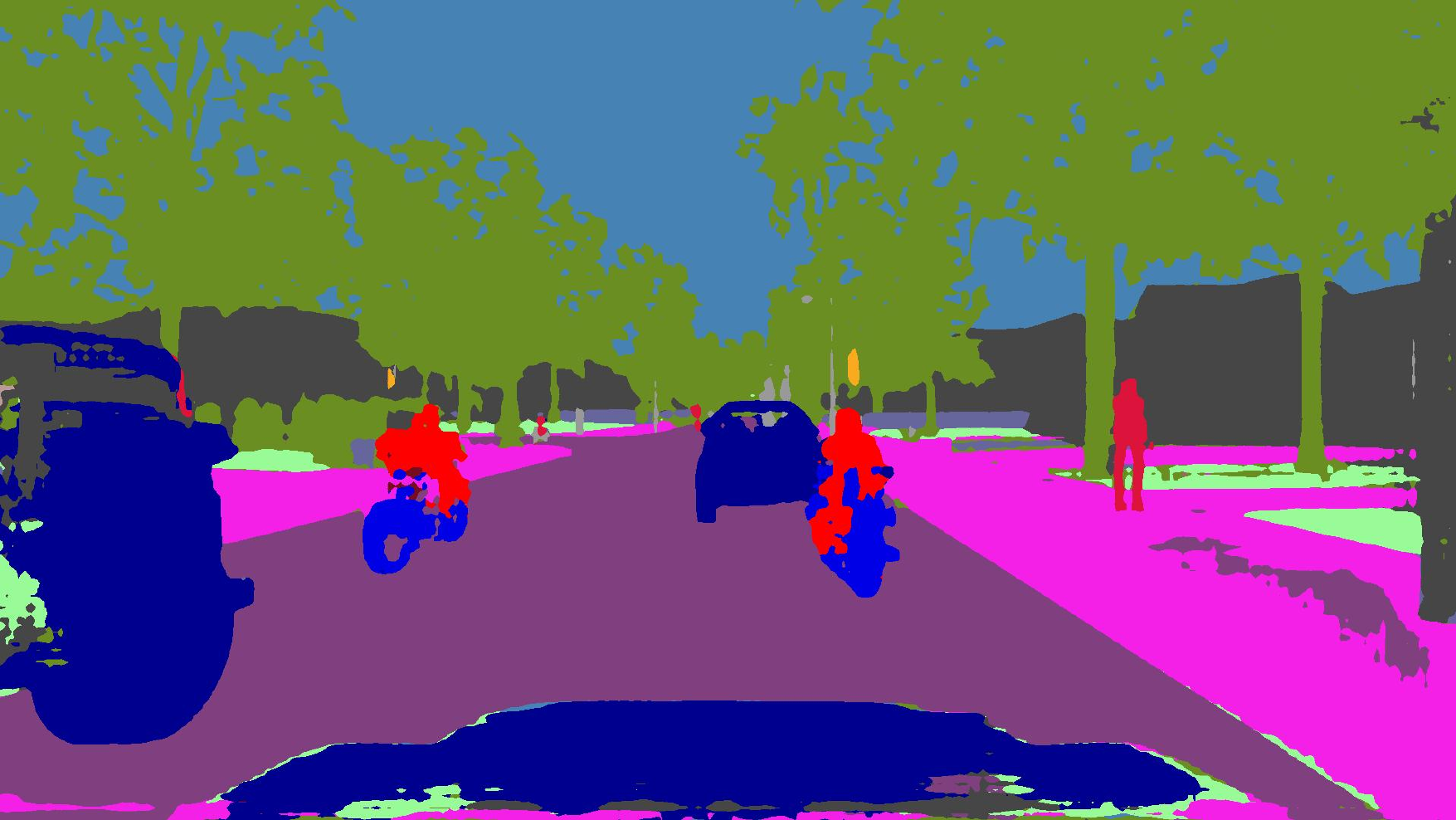}} & 
\raisebox{-0.5\height}{\includegraphics[width=29.5mm]{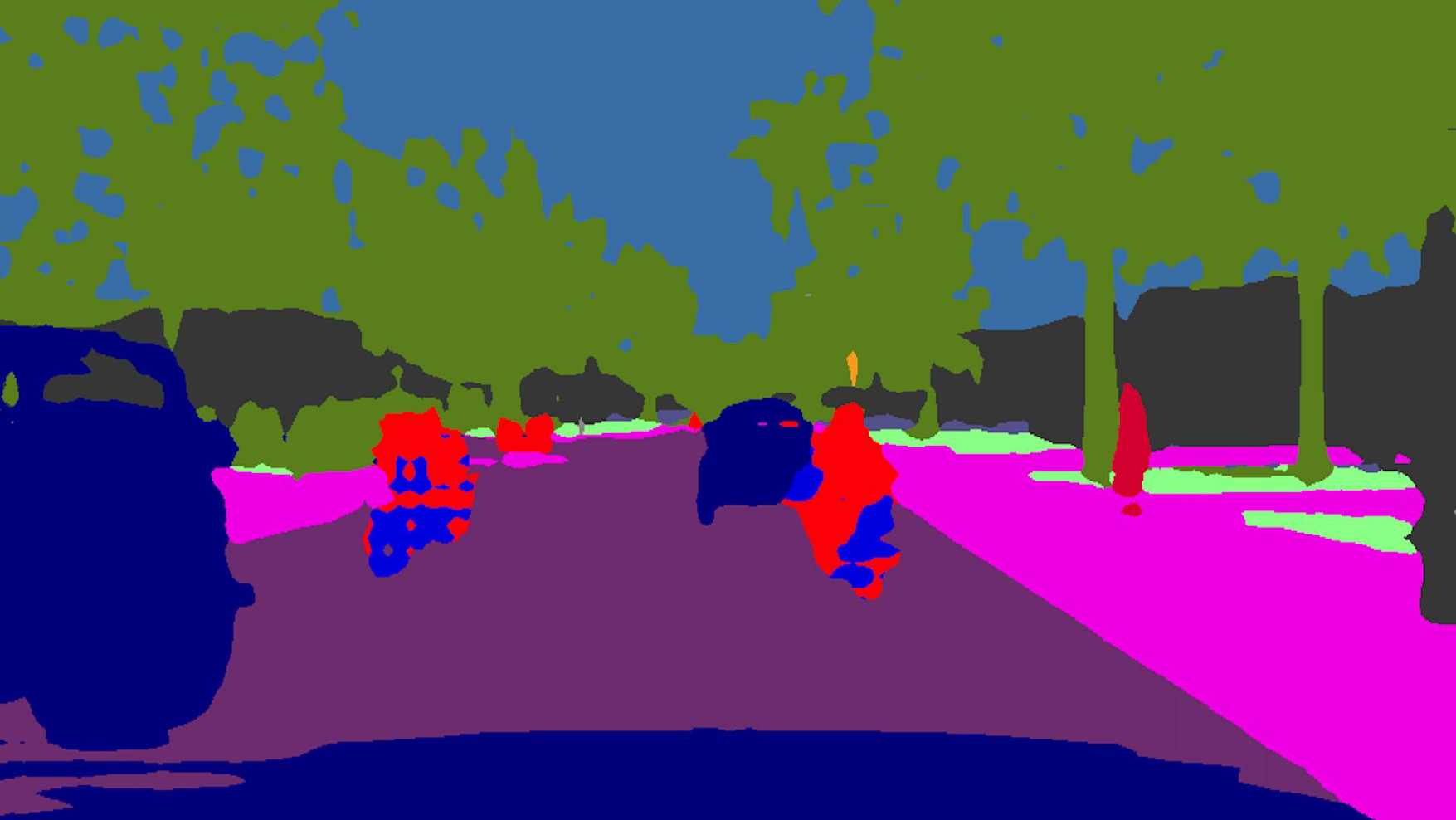}} & 
\raisebox{-0.5\height}{\includegraphics[width=29.5mm]{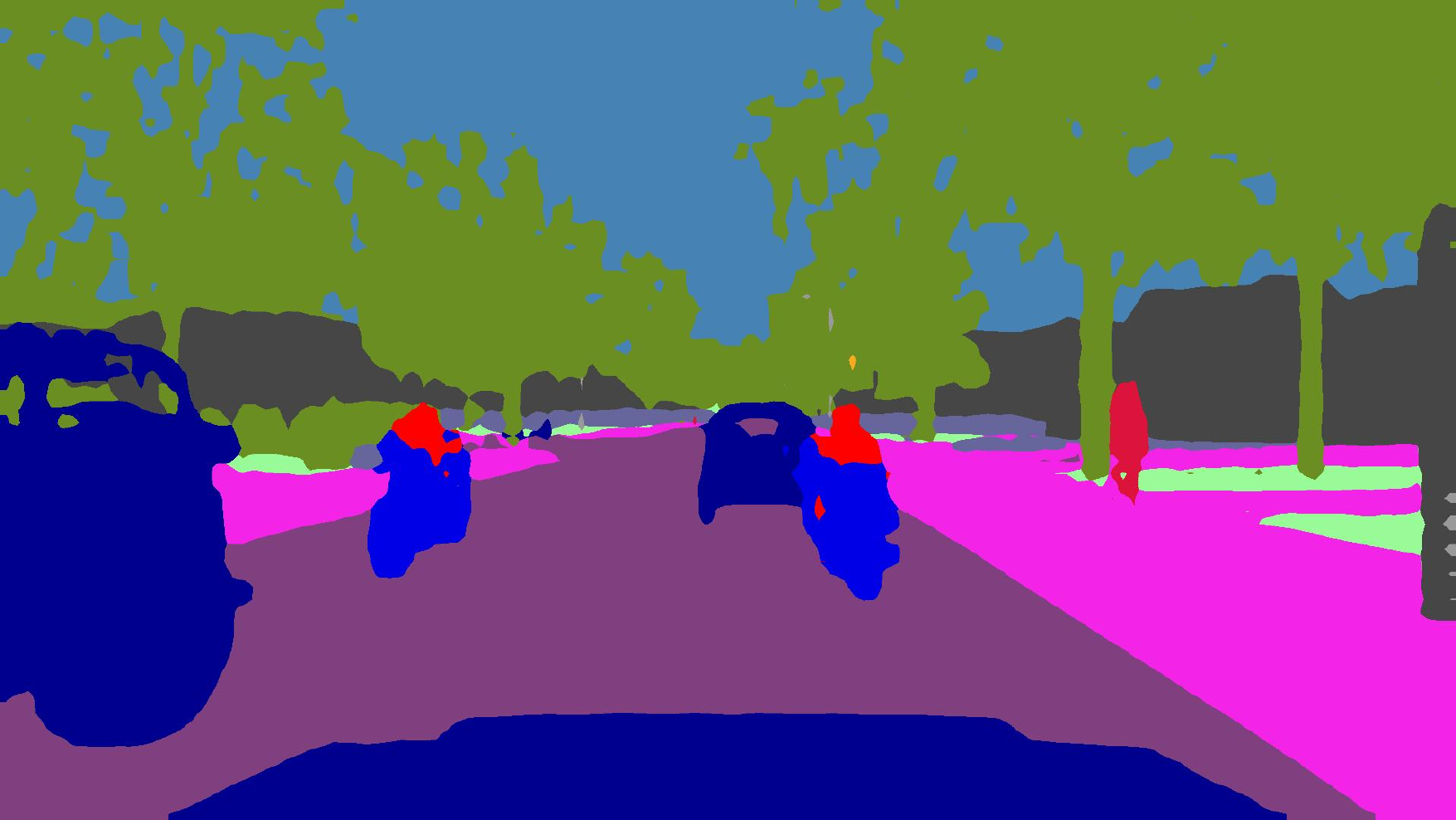}}\\&
b) Ground Truth & &                  
g) DADA & 
h) ADVENT & 
i) CLAN &  
j) DISE \\ &&&&&&
\end{tabular}
\begin{tabular}{ccccccc}
\centering
\multirow{4}{*}{\rotatebox{90}{City Change\hspace{5.5mm}}} & 
\raisebox{-0.5\height}{\includegraphics[width=29.5mm]{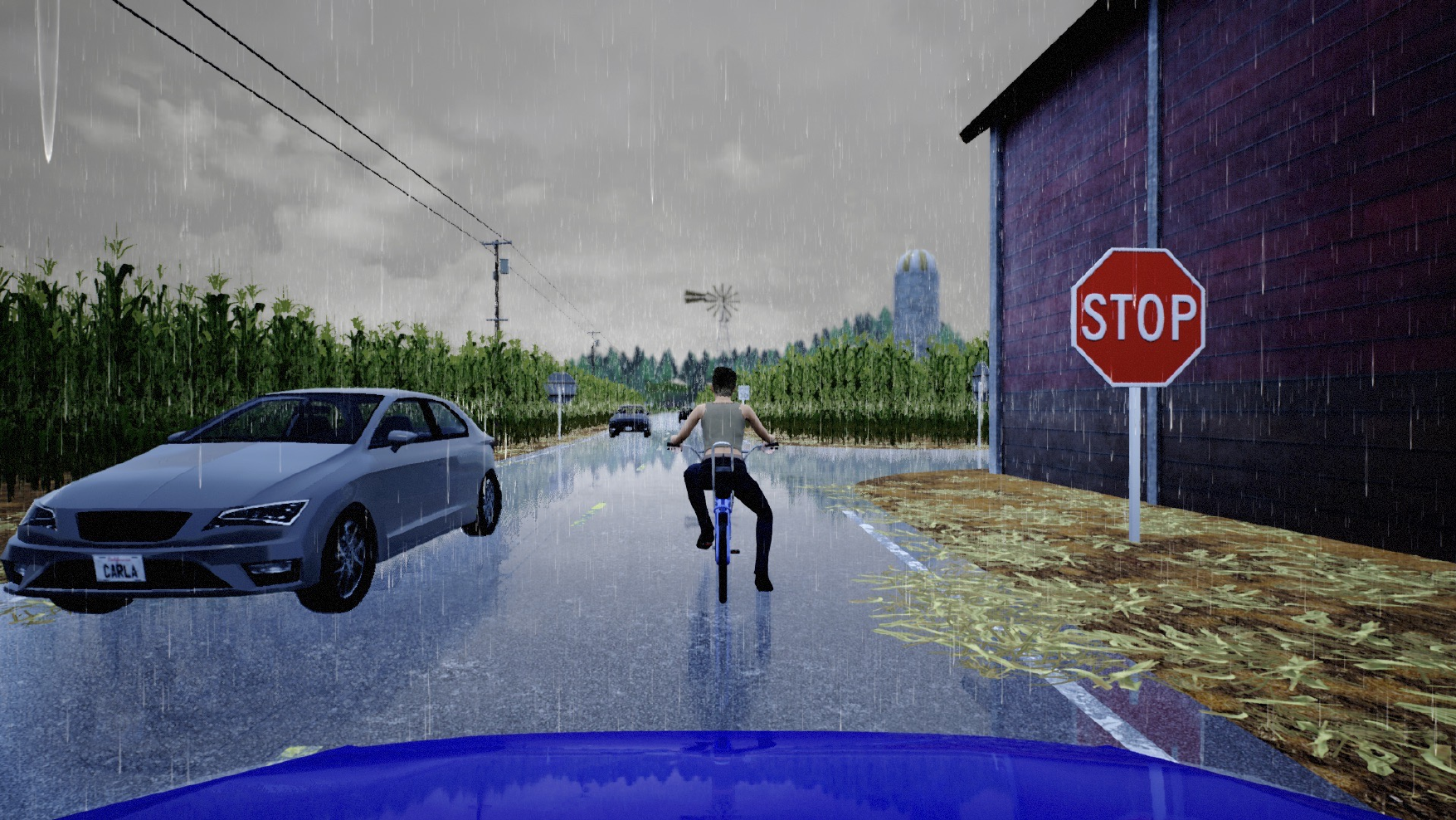}} &    &               \raisebox{-0.5\height}{\includegraphics[width=29.5mm]{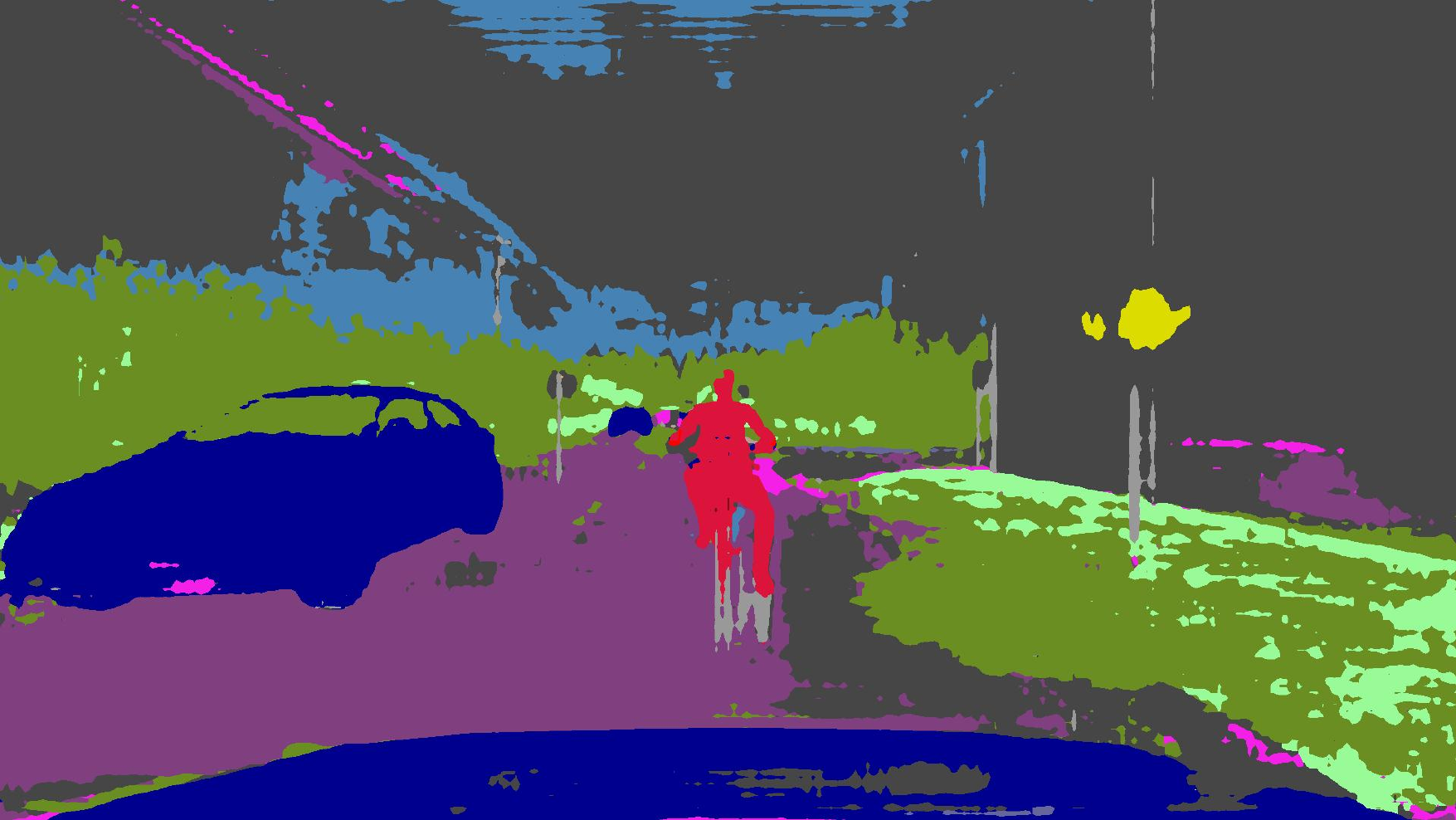}} & 
\raisebox{-0.5\height}{\includegraphics[width=29.5mm]{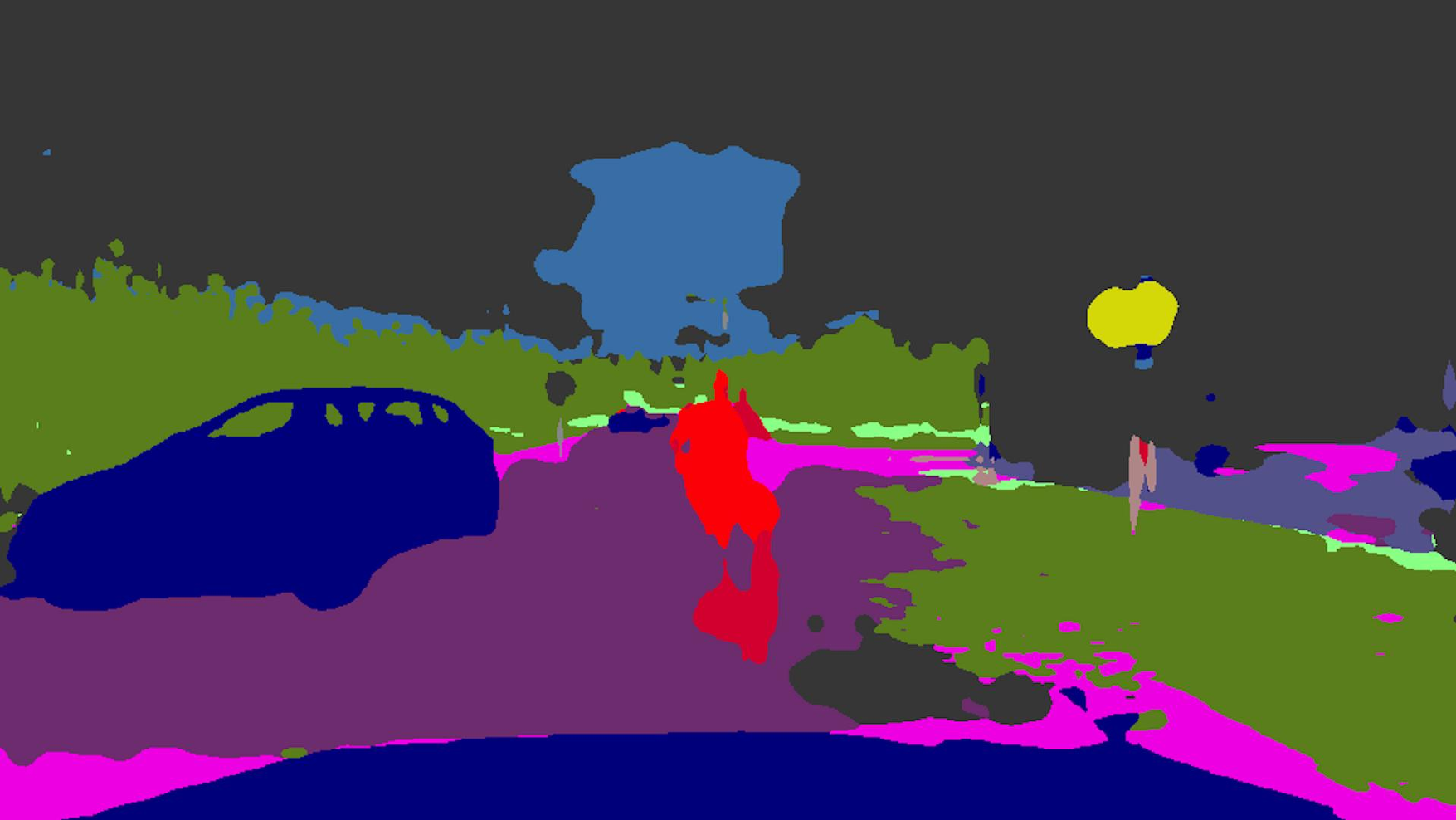}} & 
\raisebox{-0.5\height}{\includegraphics[width=29.5mm]{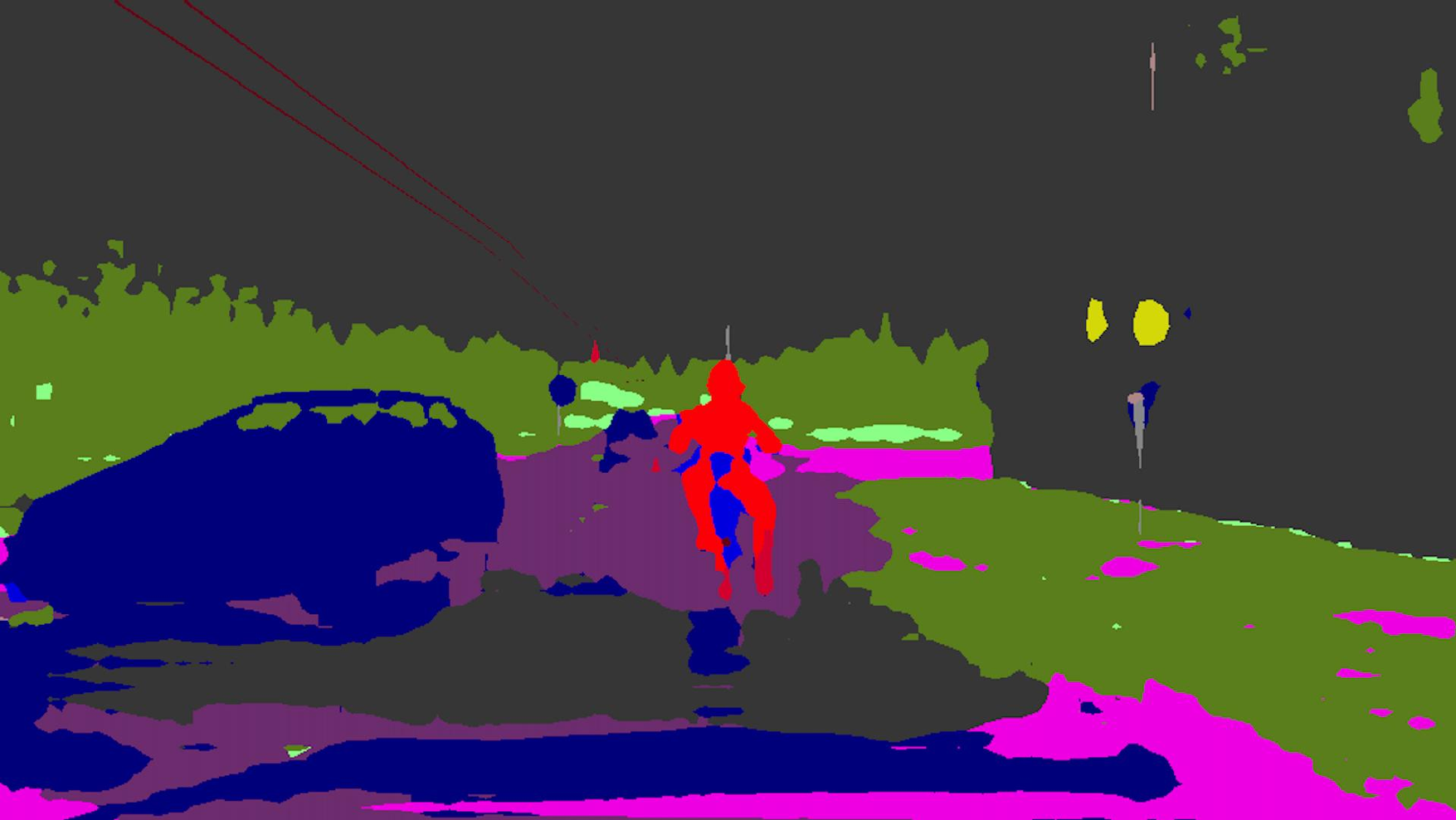}} &  
\raisebox{-0.5\height}{\includegraphics[width=29.5mm]{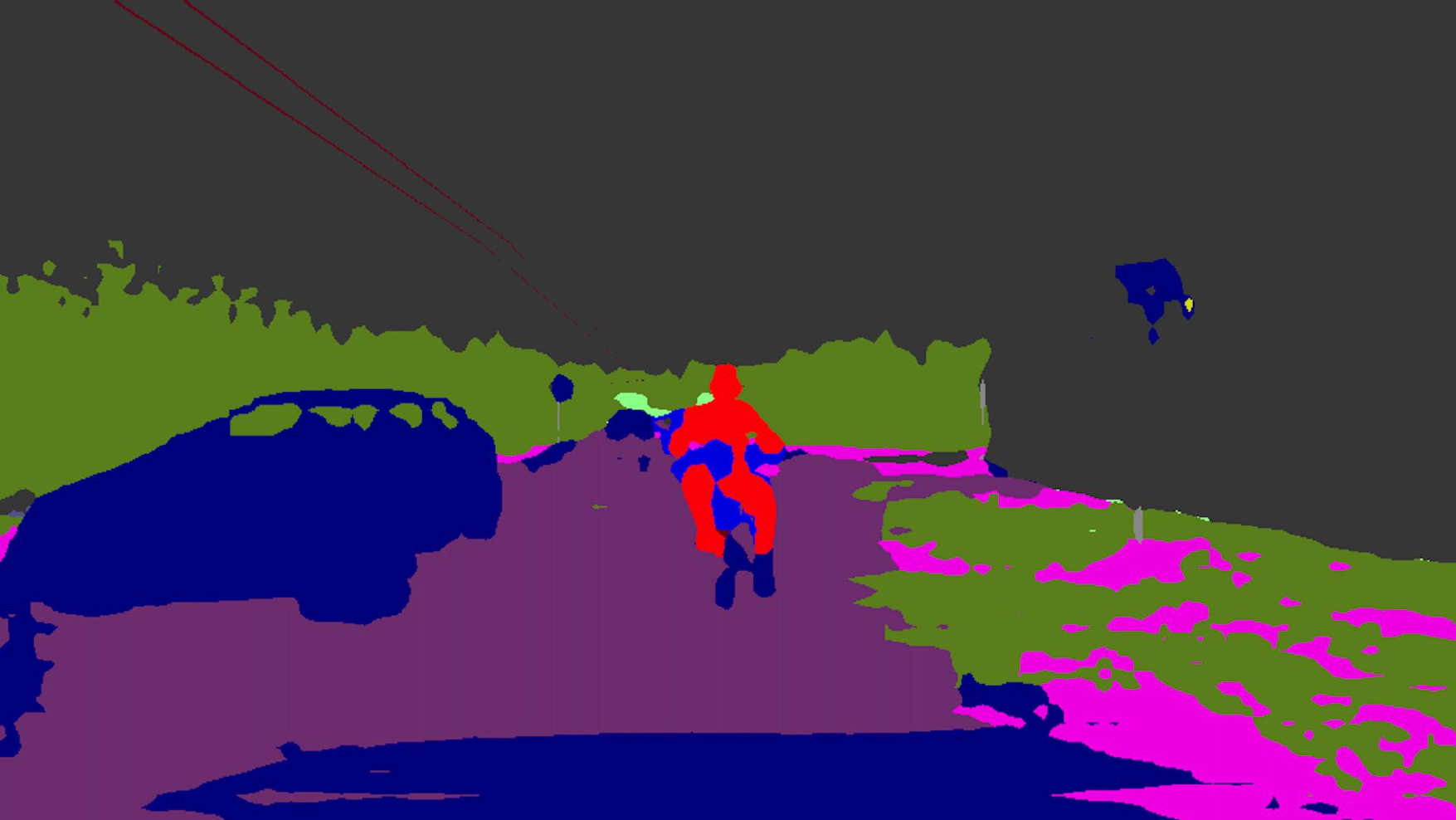}}\\&
k) RGB &       &            
m) DeepLab V2 & 
n) DeepLab V3+ & 
o) PSPNet &  
p) PSANet\\&
\raisebox{-0.5\height}{\includegraphics[width=29.5mm]{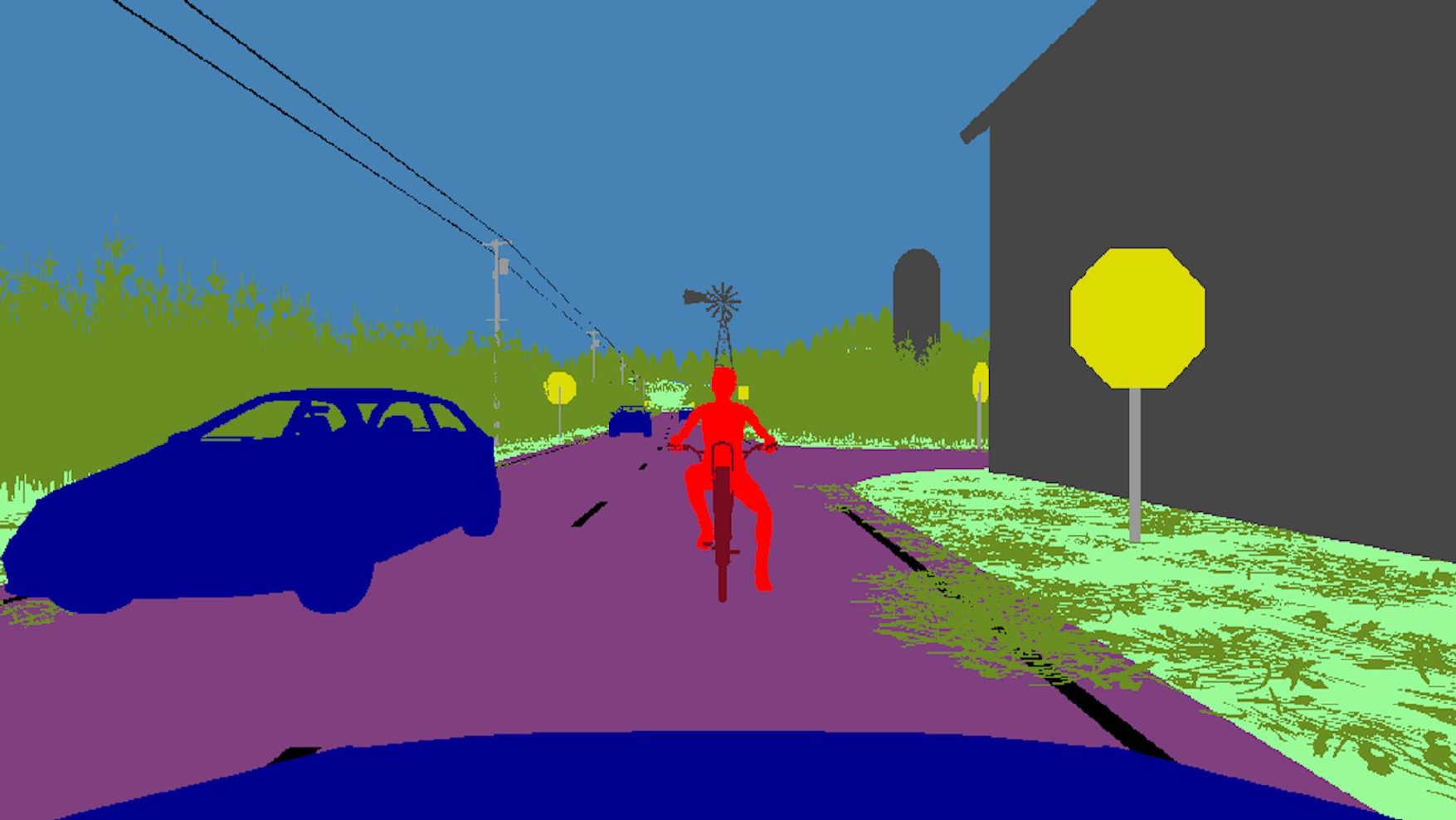}} & &
\raisebox{-0.5\height}{\includegraphics[width=29.5mm]{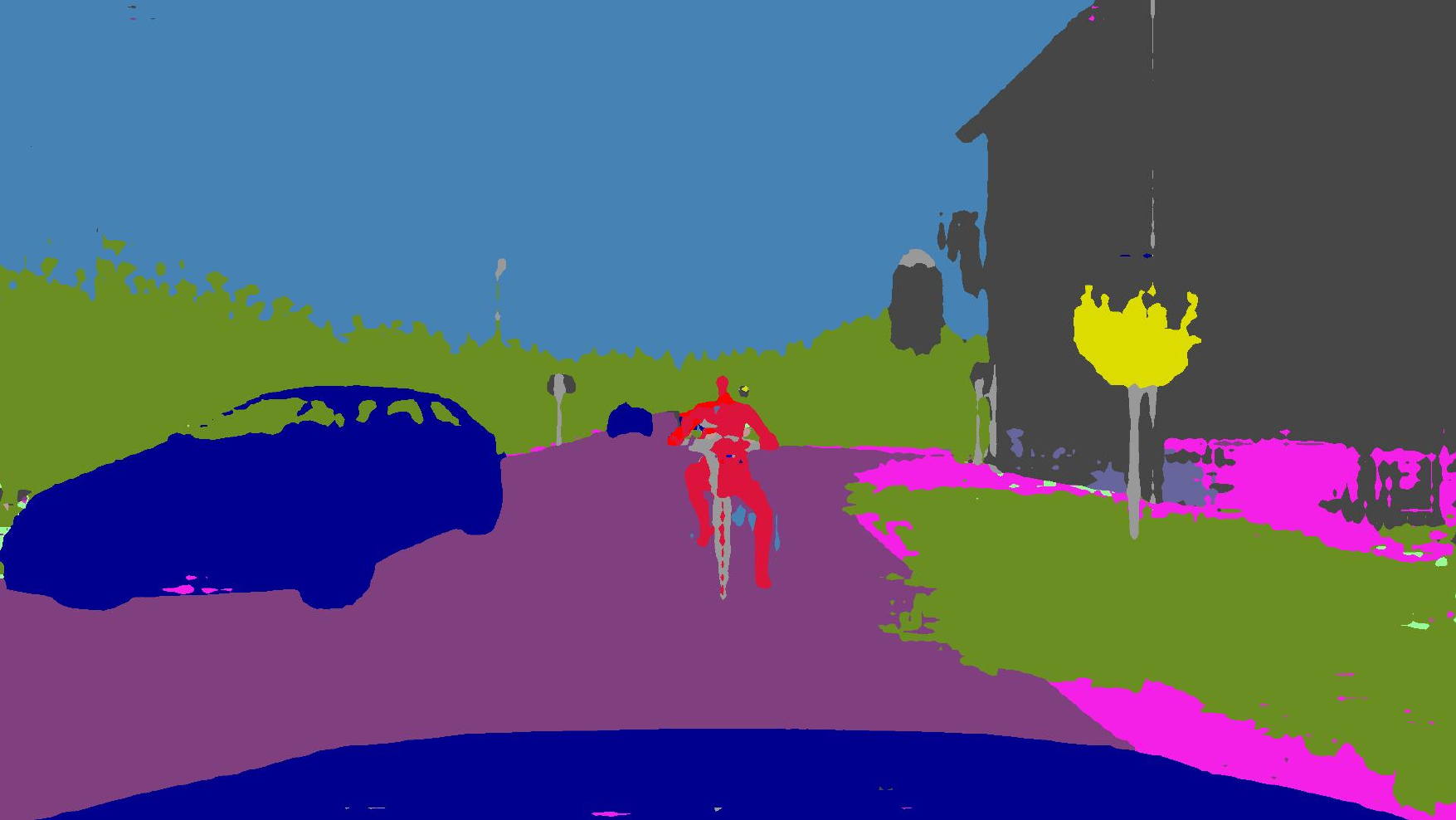}} &
\raisebox{-0.5\height}{\includegraphics[width=29.5mm]{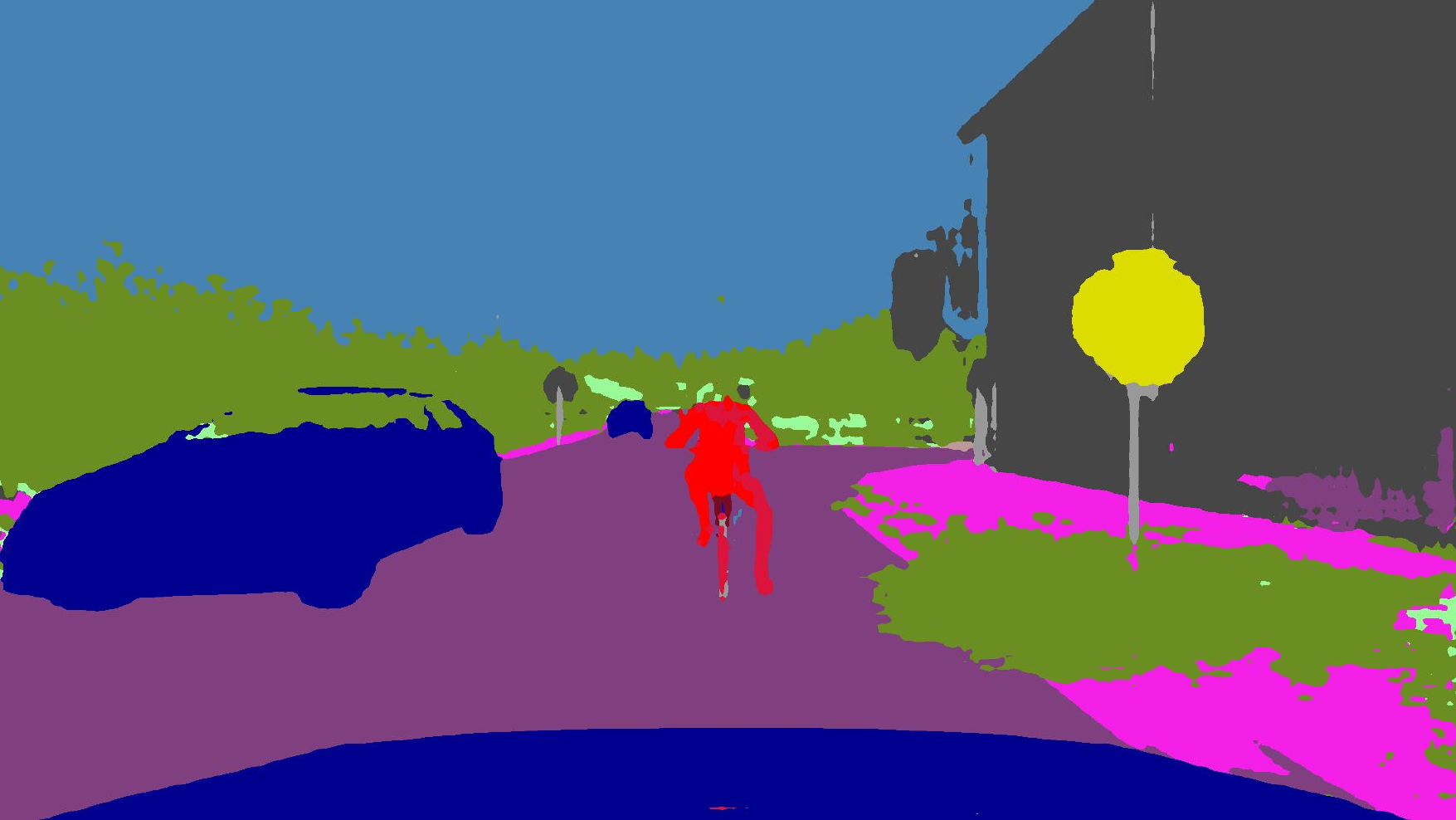}} & 
\raisebox{-0.5\height}{\includegraphics[width=29.5mm]{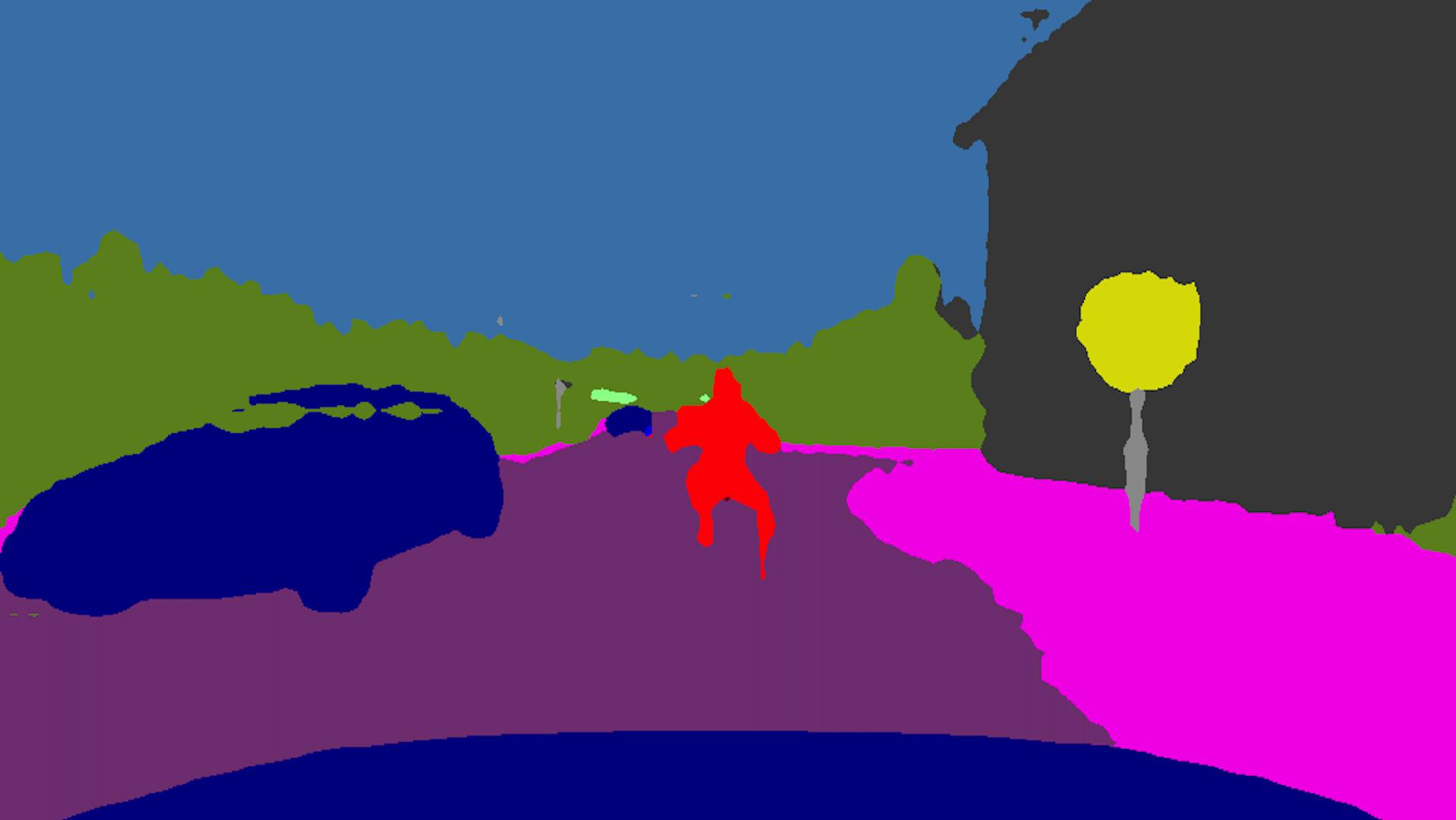}} & 
\raisebox{-0.5\height}{\includegraphics[width=29.5mm]{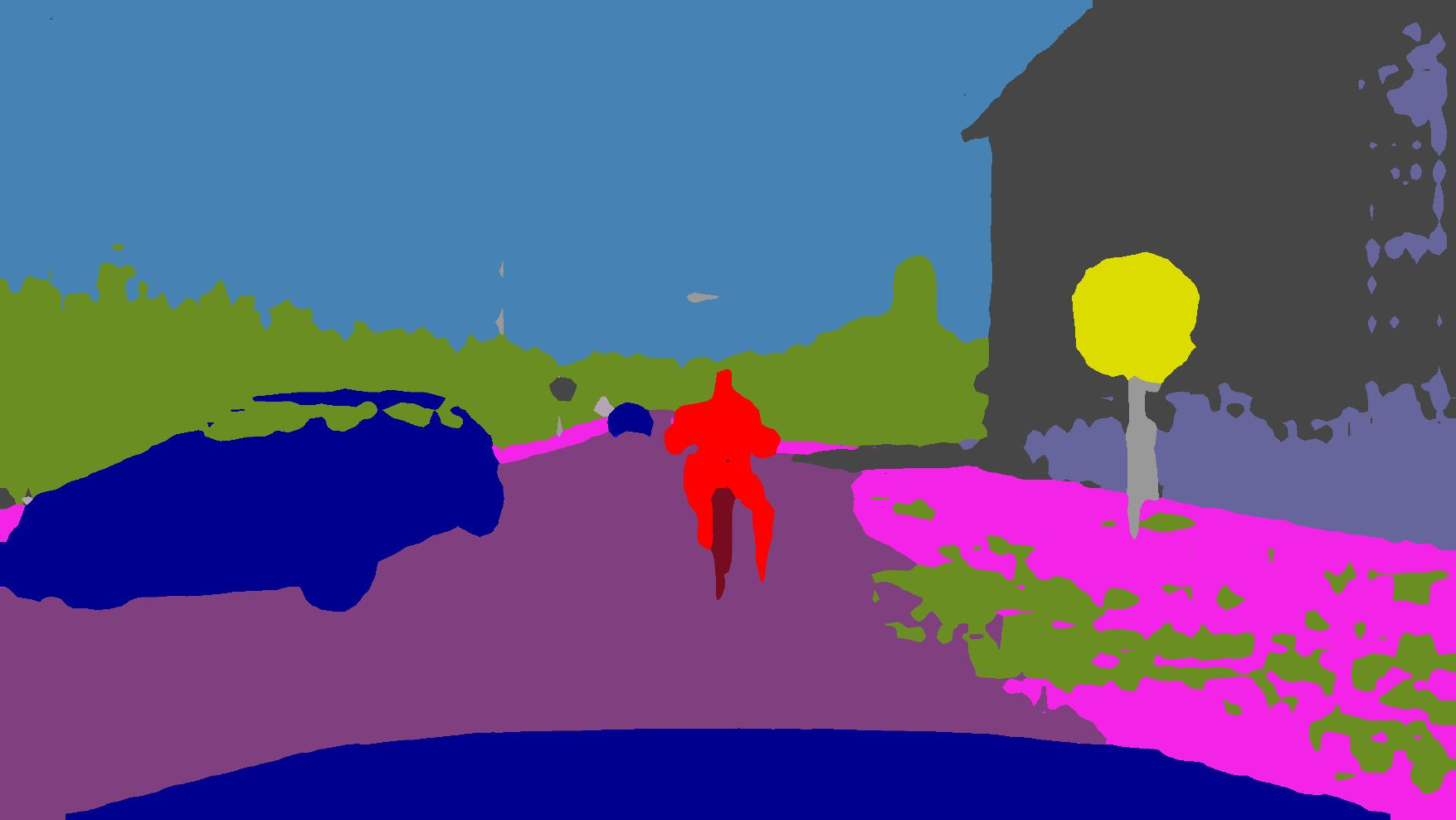}}\\&
l) Ground Truth &    &               
q) DADA & 
r) ADVENT & 
s) CLAN &  
t) DISE 
\end{tabular}
\caption{Qualitative results for the viewpoint and background change experiment. Note the more severe side effect caused by the domain shift across different cities.}
\label{fig:ScenariosExpImages}
\end{figure*}

To better compare with Cityscapes, since it is the main real dataset for benchmarking SemSeg for autonomous driving, 
we excluded from our experiments those classes that were either ambiguous (dynamic, static, other) or not present in the reference dataset (road line). 
We ended up considering the 16 labels in Fig. \ref{fig:pixeldistr}.

To quantify the distance between source and target domain (similar to 3.1 in \cite{best_practices}) we extract, using ResNet-101 \cite{resnet} pretrained on ImageNet, the features of the first 500 samples of each domain and we reduce the dimensionality (using PCA) taking the first 50 principal components. Then we proceed in two directions: in one case we compute the mean-feature vector for each domain and we measure the Euclidean and the Cosine distances, in the other case we compute the feature-wise Bhattacharyya distance.

Lastly, in all the experiments the performance is measured using the mean Intersection over Union (mIoU) metric.

\subsection{Assessing IDDA}
\label{subsection:assessing_idda}
We test the ability of the selected networks to adapt to a new domain by considering three cases that cover the three variability factors:
\begin{itemize}
    \item the first, tests the viewpoint change (from A as source to J as target), fixing background and weather (T01, CS);
    \item the second, tests the weather shifting (from CS as source to HRN as target), fixing viewpoint and background (J, T01)
    \item the third, considers two scenarios that take place in different environments (T01 as source and T07 as target), fixing viewpoint and weather (A, HRN); 
\end{itemize}
\begin{table}[ht]
\caption{Assessing IDDA Experiment Results}
\begin{adjustbox}{width=1.0\columnwidth}
\begin{tabular}{@{}llccc@{}}
\toprule
\multicolumn{2}{l}{\multirow{2}{*}{Semantic Segmentation Networks}}                                        & \multicolumn{3}{c}{Scenarios (\% mIoU)}                                                                                                                                                                 \\ \cmidrule(l){3-5} 
\multicolumn{2}{l}{}                                                                                       & Viewpoint Change                                            & Weather Change                                                   & City Change                                                   \\ \cmidrule(l){3-5} 
                        & \multicolumn{1}{r}{\begin{tabular}[c]{@{}r@{}}Source: \\ Target:\end{tabular}}   & \begin{tabular}[c]{@{}c@{}}T01 CS A\\ T01 CS J\end{tabular} & \begin{tabular}[c]{@{}c@{}}T01 CS J\\ T01 HRN J\end{tabular} & \begin{tabular}[c]{@{}c@{}}T01 HRN A\\ T07 HRN A\end{tabular} \\ \midrule
\multirow{5}{*}{w/o DA} & DeepLab V2                                                                       & 62.60                                                       & 40.24                                                           & 21.65                                                         \\ \cmidrule(l){2-5} 
                        & DeepLab V3+                                                                      & 64.93                                                       & 33.93                                                            & 14.27                                                         \\ \cmidrule(l){2-5} 
                        & PSPNet                                                                           & 67.32                                                       & 29.65                                                            & 14.64                                                         \\ \cmidrule(l){2-5} 
                        & PSANet                                                                           & 66.88                                                       & 33.60                                                            & 15.52                                                         \\ \cmidrule(l){2-5} 
                        & DeepLab V2 (source=target)& 79.13                                              & 78.31                                                            & 78.02                                                \\ \midrule
\multirow{4}{*}{w/ DA}  & DADA                                                                             & 66.42                                                       & 55.87                                                            & 36.48                                                         \\ \cmidrule(l){2-5} 
                        & ADVENT                                                                           & 68.43                                                       & 61.13                                                            & 39.30                                                         \\ \cmidrule(l){2-5} 
                        & CLAN                                                                             & 70.30                                                       & 65.52                                                            & 41.18                                                         \\ \cmidrule(l){2-5} 
                        & DISE                                                                             & 73.64                                                       & 71.91                                                            & 46.71                                                         \\ \bottomrule
\end{tabular}
\end{adjustbox}
\label{table:ScenarioExperiments}
\end{table}
We used the method detailed in the section \ref{section:setup} to measure numerically the distance and the difficulty of the three cases (see Tab. \ref{table:scenariosDistance}). As a visual confirmation, we use tSNE to project the features extracted with the ResNet-101 into a more comprehensible 2D space (see Fig. \ref{fig:tSNE4Scenarios}). 

The results of the experiments are reported in Tab. \ref{table:ScenarioExperiments}. As expected from the distance computation, the shift across cities, made even more challenging by the rainy condition, produces a higher degradation in performance than the other two experiments.
In the town shift the SemSeg networks struggle to correctly classify the scene and their accuracy drops as low as 14\%. In this case DA produces a considerable boost with an averaged accuracy of 40\%. This trend is repeated within the shift across weathers but, since the gap among source and target domain is smaller, the resulting average mIoU is of 34\%. In this case DA performs quite well, giving as outcome an averaged 63\%. Lastly, the viewpoint change proves to be the best performing set of experiments, so the addition of DA increases the average accuracy by only 4\%. Among all of the DA networks, DISE proves to be the most capable while the depth information exploited by DADA does not seem to improve the performance. 

Fig. \ref{fig:ScenariosExpImages} illustrates some qualitative results of our experiments. Looking at the output produced we can highlight two interesting problems that seem to affect the SemSeg networks and their generalization capability. Considering the viewpoint change, all the SemSeg models without DA struggle to  classify well the portion of the image occupied by the hood of the vehicle, improperly classifying it as a building. Moreover, when changing the scenario and moving to a countryside scene with vegetation in place of roadside and sidewalks (\enquote{city change} case), we observe that, during training, all the networks (with the only exception of DeepLab V2) learned and memorized the pattern \enquote{building-sidewalk-road} of the source scenario. Therefore, when moving to the target environment they are not able to adapt and tend to incorrectly classify the terrain as sidewalk.

The diversity of our dataset and the possibility to simulate various kind of real scenario has made it possible to gain this kind of insight. Furthermore, we have demonstrated the limitations of the actual state-of-the-art SemSeg networks and how IDDA could be a powerful tool to validate the adaptation performances to a domain shift in driving applications.

\subsection{Synthetic vs. real scenarios}
\label{subsect:synvsreal}
In the second experiment we test how well the networks trained on a synthetic dataset can adapt to a real one. 
In particular, we consider two cases, each using a source domain obtained from a combination of several scenarios in IDDA:
\begin{table}[th]
%\bigskip
\caption{Distances between IDDA and real datasets}
\begin{adjustbox}{width=1.0\columnwidth}
\begin{tabular}{@{}llcccc@{}}
\toprule
\multirow{2}{*}{}                                                      & \multirow{2}{*}{\begin{tabular}[c]{@{}l@{}}Distance\\ Function\end{tabular}}      & \multicolumn{4}{c}{Dataset}                                                               \\ \cmidrule(l){3-6} 
                                                                       &                                                                              & Cityscapes & BDD100K & Mapillary & A2D2 \\ \midrule
\multirow{4}{*}{\begin{tabular}[c]{@{}l@{}}Best \\ Case\end{tabular}}  & Euclidean                                                                   & 7.4419                         & 7.6177                      & 5.4493  &  6.3874                     \\ \cmidrule(l){2-6} 
                                                                       & Cosine                 & 1.3582                         & 1.6209                      & 1.2924  & 1.0589                      \\ \cmidrule(l){2-6} 
                                                                       & Bhattacharyya                                                      & 0.0552                         & 0.0502                      & 0.0106 & 0.0447                        \\ \midrule
\multirow{4}{*}{\begin{tabular}[c]{@{}l@{}}Worst \\ Case\end{tabular}} & Euclidean                                                                    & 8.2360                        & 7.7618                      & 4.9548 & 7.0150                        \\ \cmidrule(l){2-6} 
                                                                       & Cosine                                           & 1.5465                         & 1.5526                      & 0.9147 & 1.1849                        \\ \cmidrule(l){2-6} 
                                                                       & Bhattacharyya                                                               & 0.0498                         & 0.0381                      & 0.0267 & 0.0387                        \\ \bottomrule 
\end{tabular}
\end{adjustbox}
\label{table:SynVSRealDistance}
\end{table}

\begin{figure}[htpb]
    \centering
    \includegraphics[width=1.0\columnwidth]{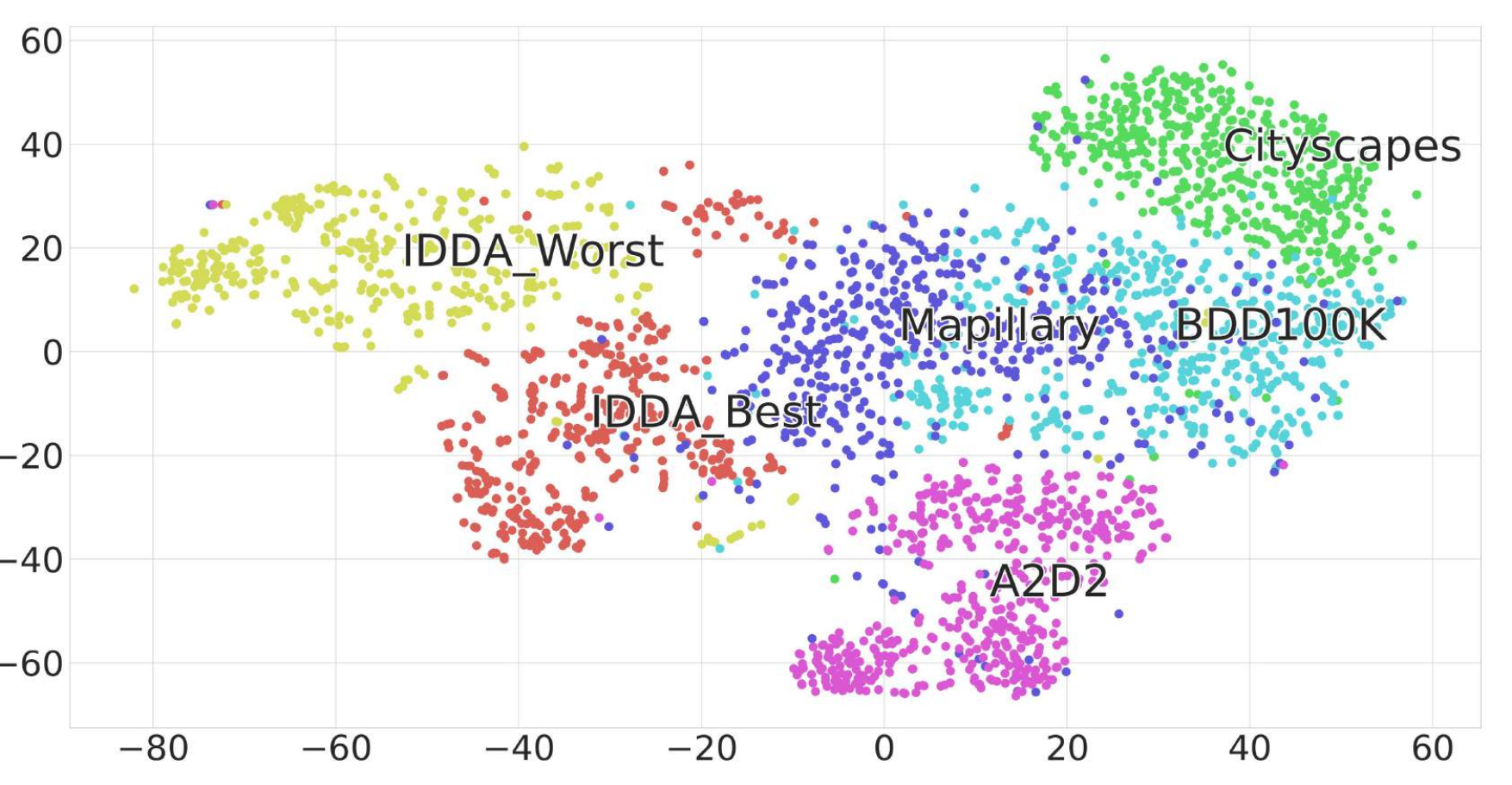}
    \vspace{-5mm}
    \caption{The tSNE representation of the distributions of synthetic and real datasets.}
    \label{fig:tsne_synvsreal}
\end{figure}

\begin{itemize}
    \item the first, called ``best case'', is a mixture of samples with similar environmental conditions to the target domains, counting a total of 29,952 
    elements sampled in a stratified fashion and taken only from urban environments (T01-T06), with a car-like point-of-view (A or M) and clear weather conditions at noon (CN);

    \item the second, called ``worst case'', has a higher visual discrepancy w.r.t. the target samples and it counts 40,128 samples taken from the previously excluded countryside town (T07), with a hooded and a non-hooded point of views (J and B) and rainy conditions at noon (HRN).
\end{itemize}

Tab. \ref{table:SynVSRealDistance} and Fig. \ref{fig:tsne_synvsreal} show numerically and visually the distance among the dataset distributions. When evaluating the performance on the target datasets, we ignored all labels not included in Fig. \ref{fig:pixeldistr} and labeled all four-wheeled vehicles as our semantic class \enquote{vehicle}.
For A2D2 we considered only 13 labels due to its labeling inconsistencies with IDDA, e.g. the complete absence of class ``rider'' and ``wall'' and the union of ``vegetation'' with ``terrain''.
Results are in Tab. \ref{table:SynVsRealExperiments}.
\begin{figure*}
%\bigskip
\centering
\begin{tabular}{cccccc}
\centering
\multirow{4}{*}{\rotatebox{90}{Best Case w/ Cityscapes\hspace{-3mm}}} & 
\raisebox{-0.5\height}{\includegraphics[width=37.5mm]{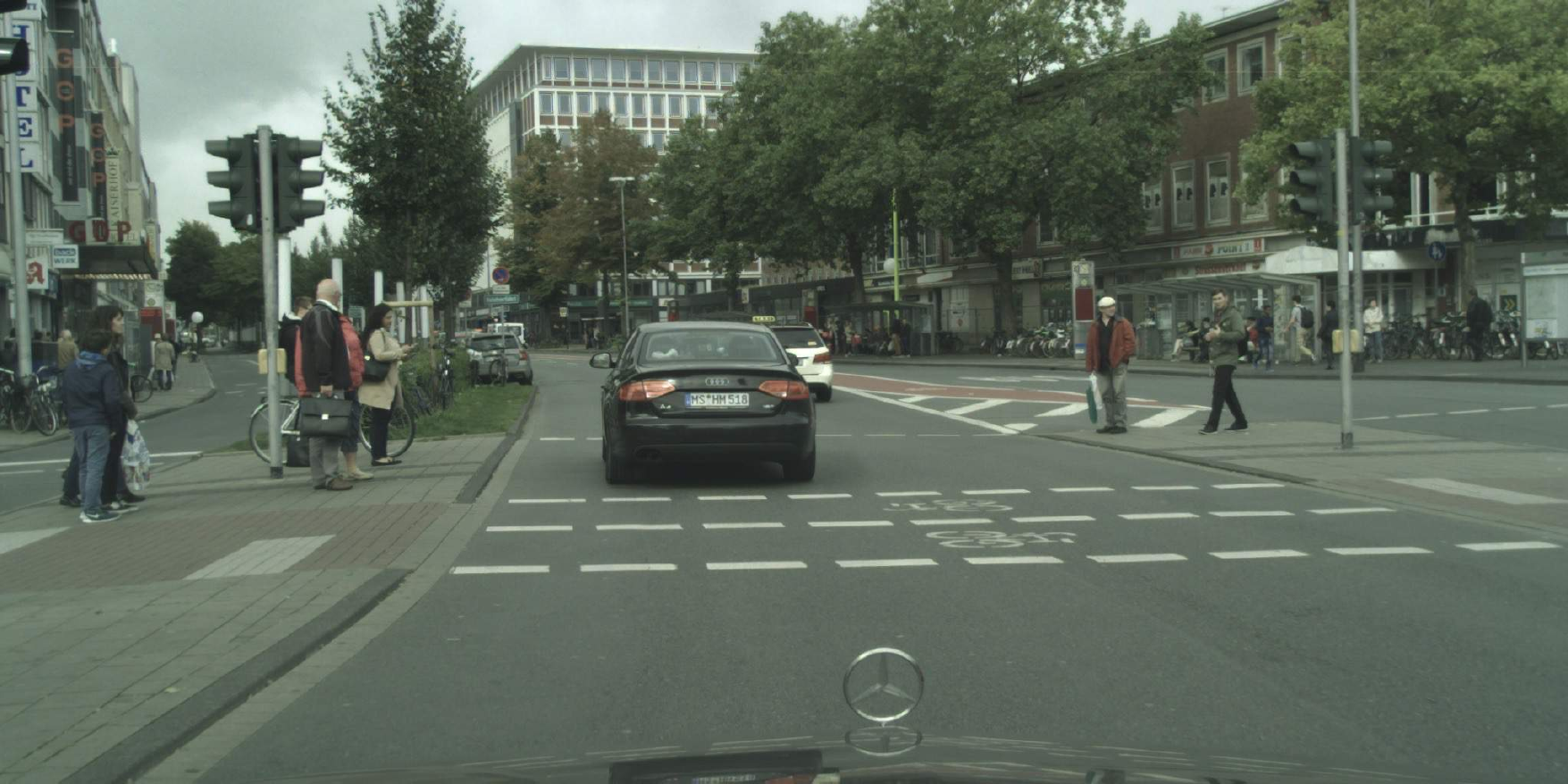}} &    &               \raisebox{-0.5\height}{\includegraphics[width=37.5mm]{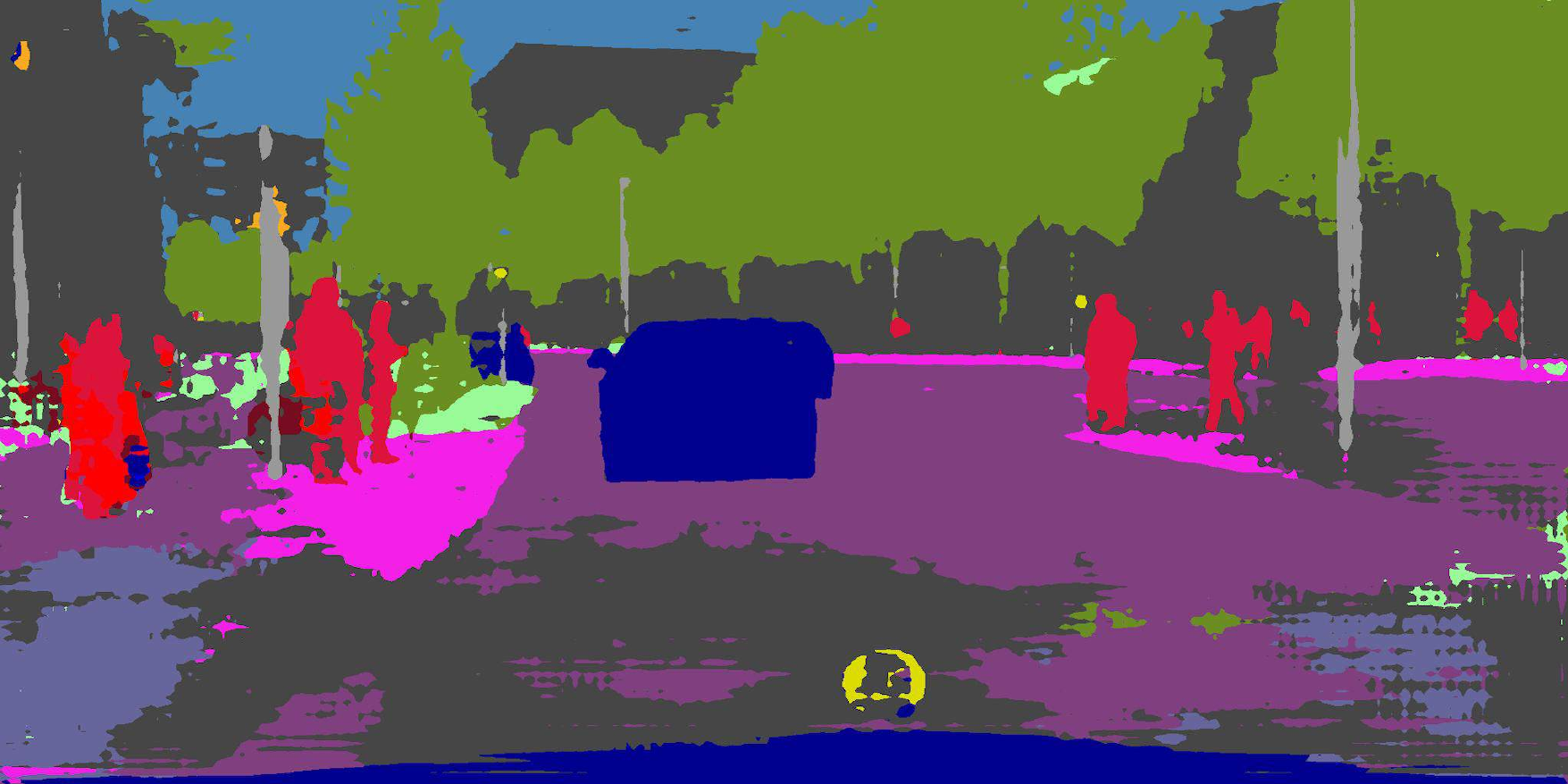}} & 
\raisebox{-0.5\height}{\includegraphics[width=37.5mm]{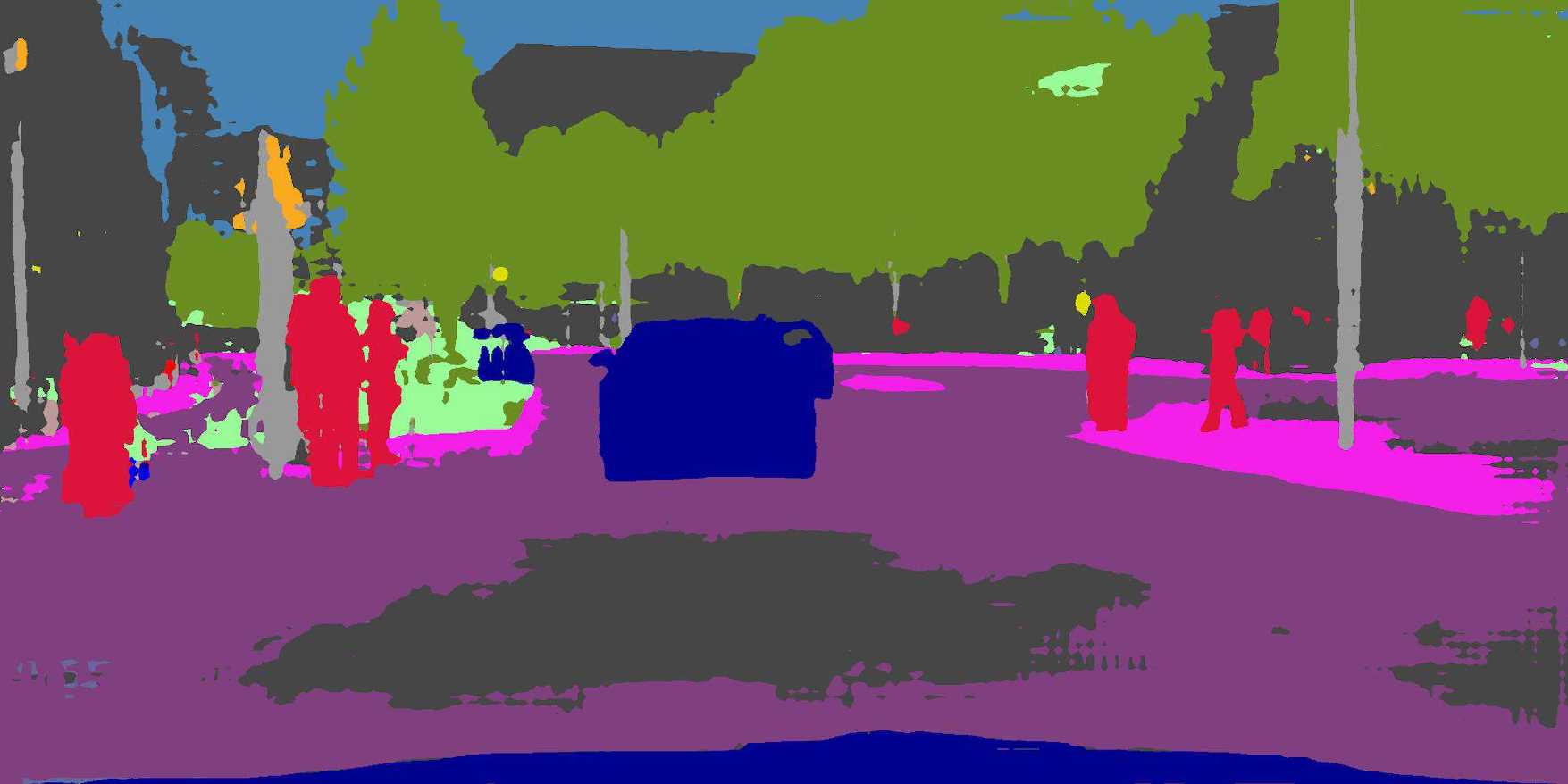}} & 
\raisebox{-0.5\height}{\includegraphics[width=37.5mm]{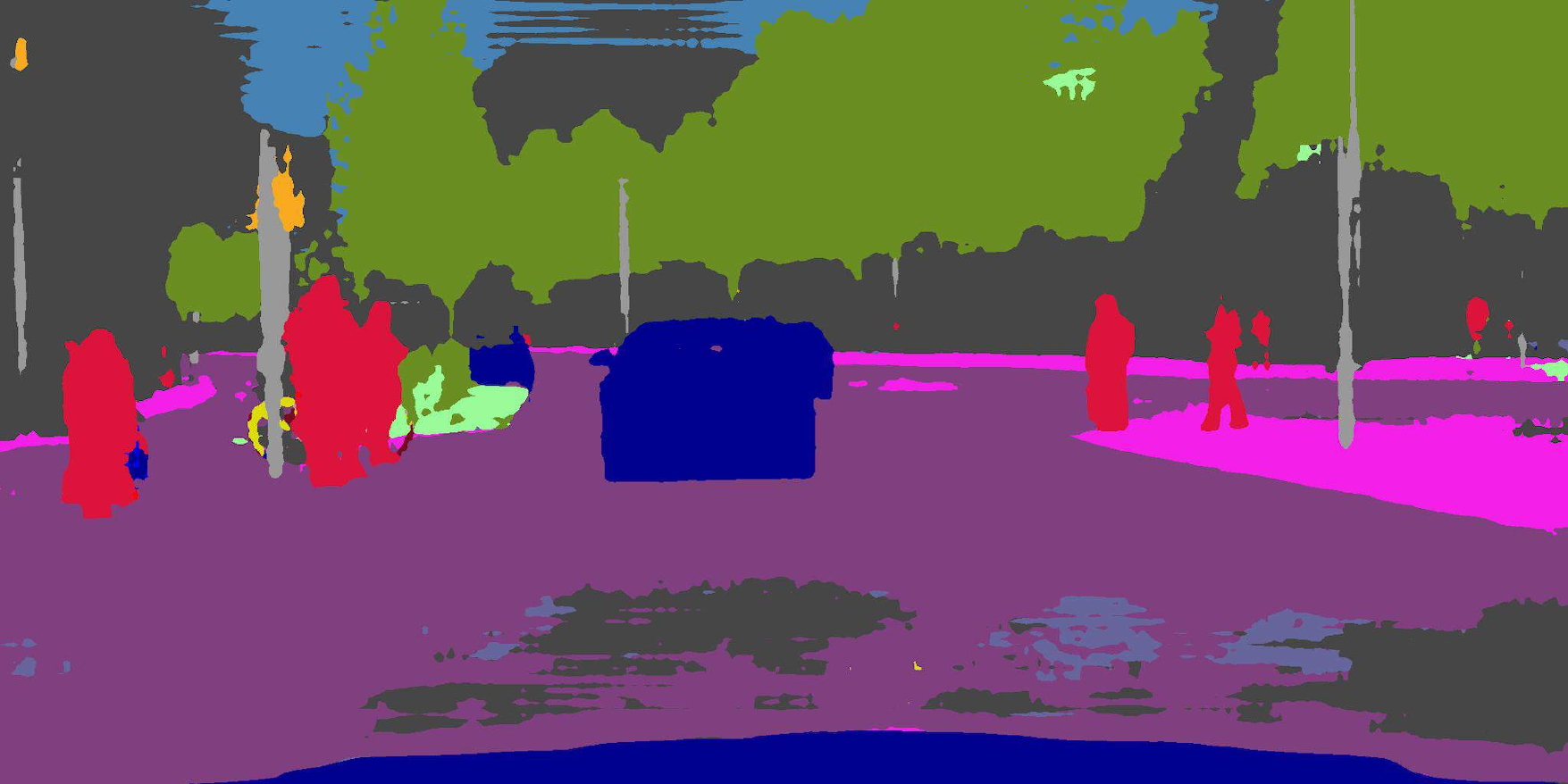}}\\&
a) RGB &   &                
c) DeepLab V2 & 
d) DADA & 
e) ADVENT\\&
\raisebox{-0.5\height}{\includegraphics[width=37.5mm]{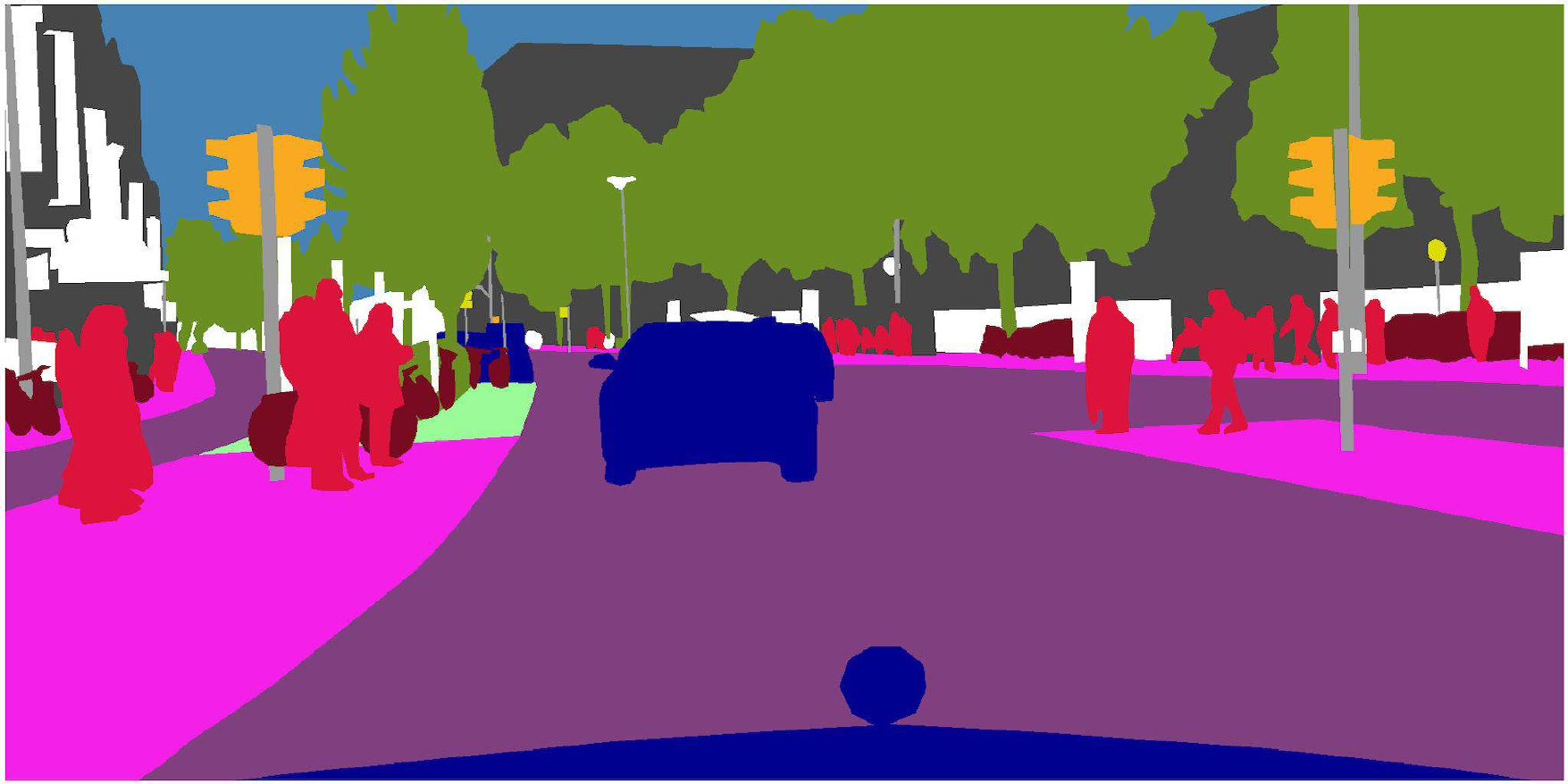}} & &
\raisebox{-0.5\height}{\includegraphics[width=37.5mm]{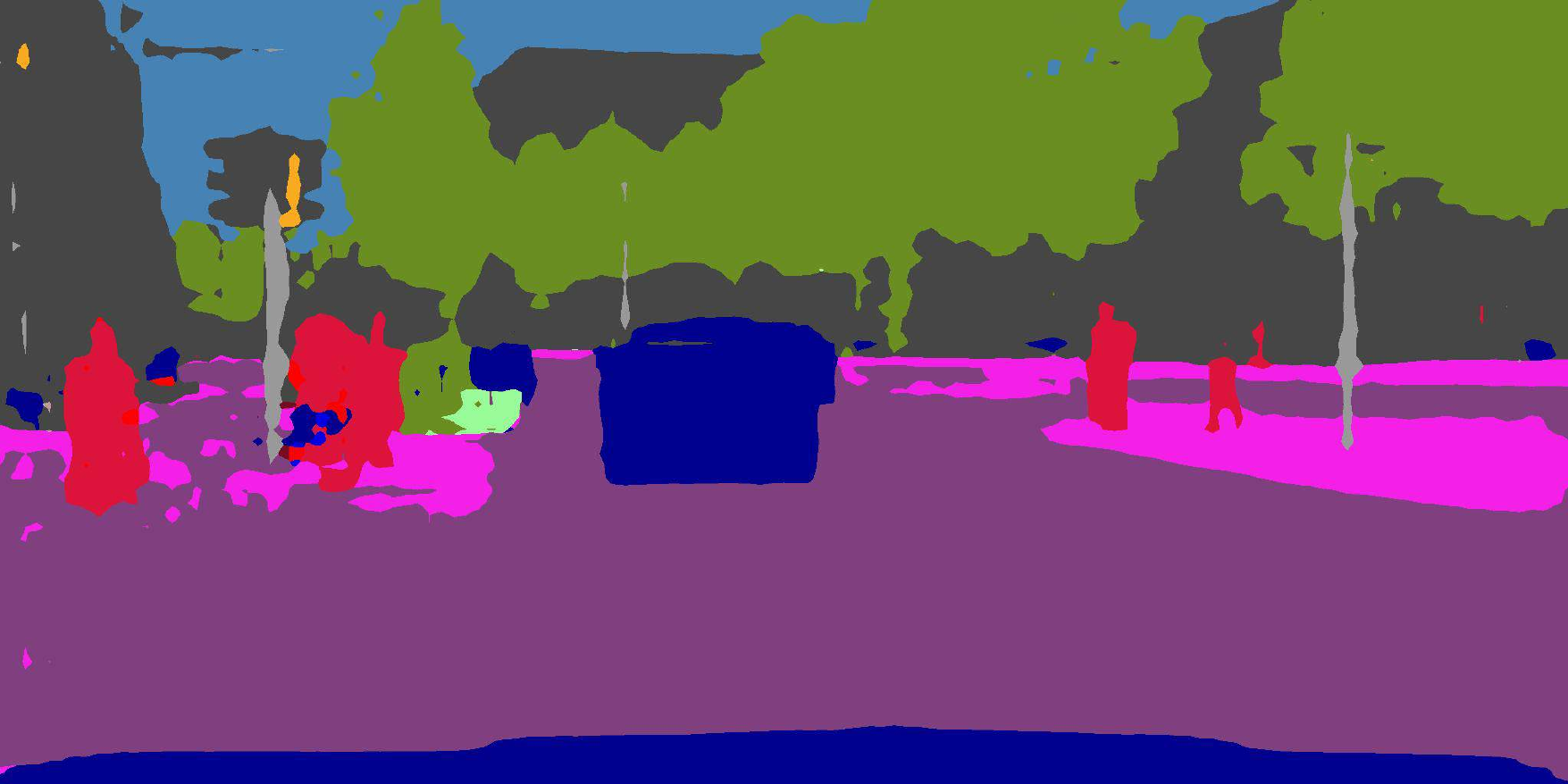}} &
\raisebox{-0.5\height}{\includegraphics[width=37.5mm]{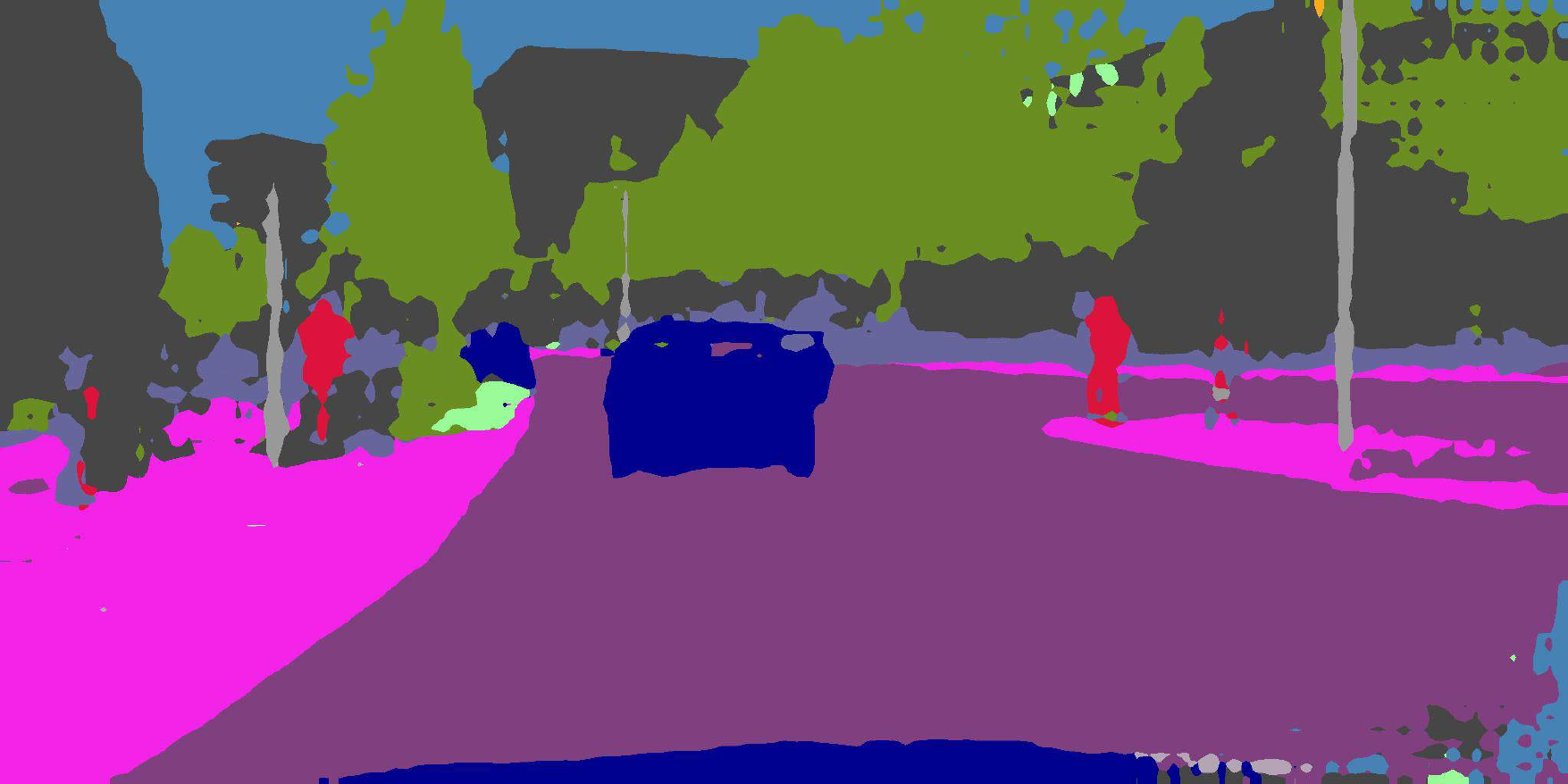}} & 
\raisebox{-0.5\height}{\includegraphics[width=37.5mm]{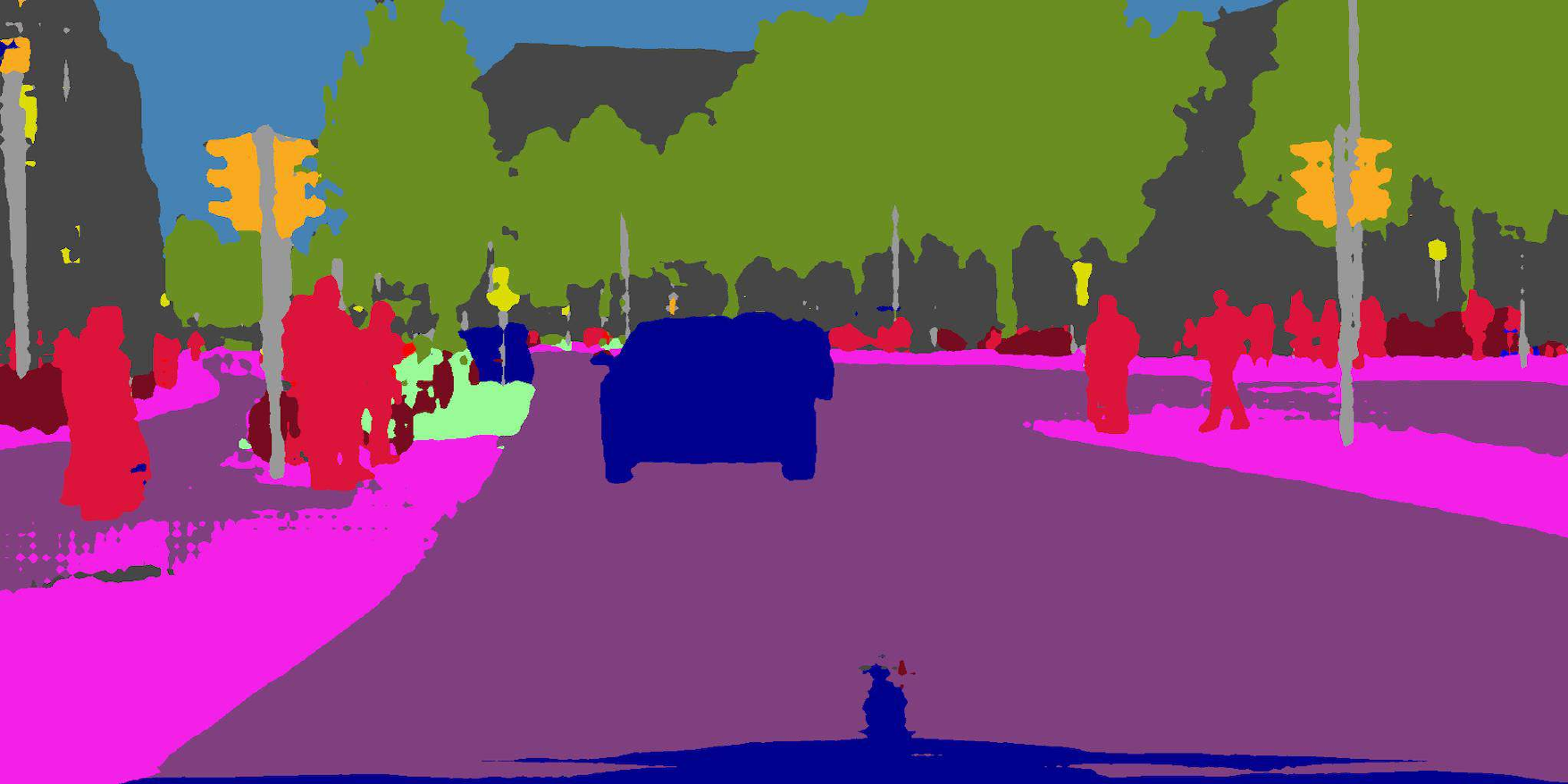}}\\&
b) Ground Truth & &                  
f) CLAN & 
g) DISE & 
h) Baseline\\ &&&&&
\end{tabular}
\begin{tabular}{cccccc}
\centering
\multirow{4}{*}{\rotatebox{90}{Worst Case w/ BDD100K\hspace{-3mm}}} & 
\raisebox{-0.5\height}{\includegraphics[width=37.5mm]{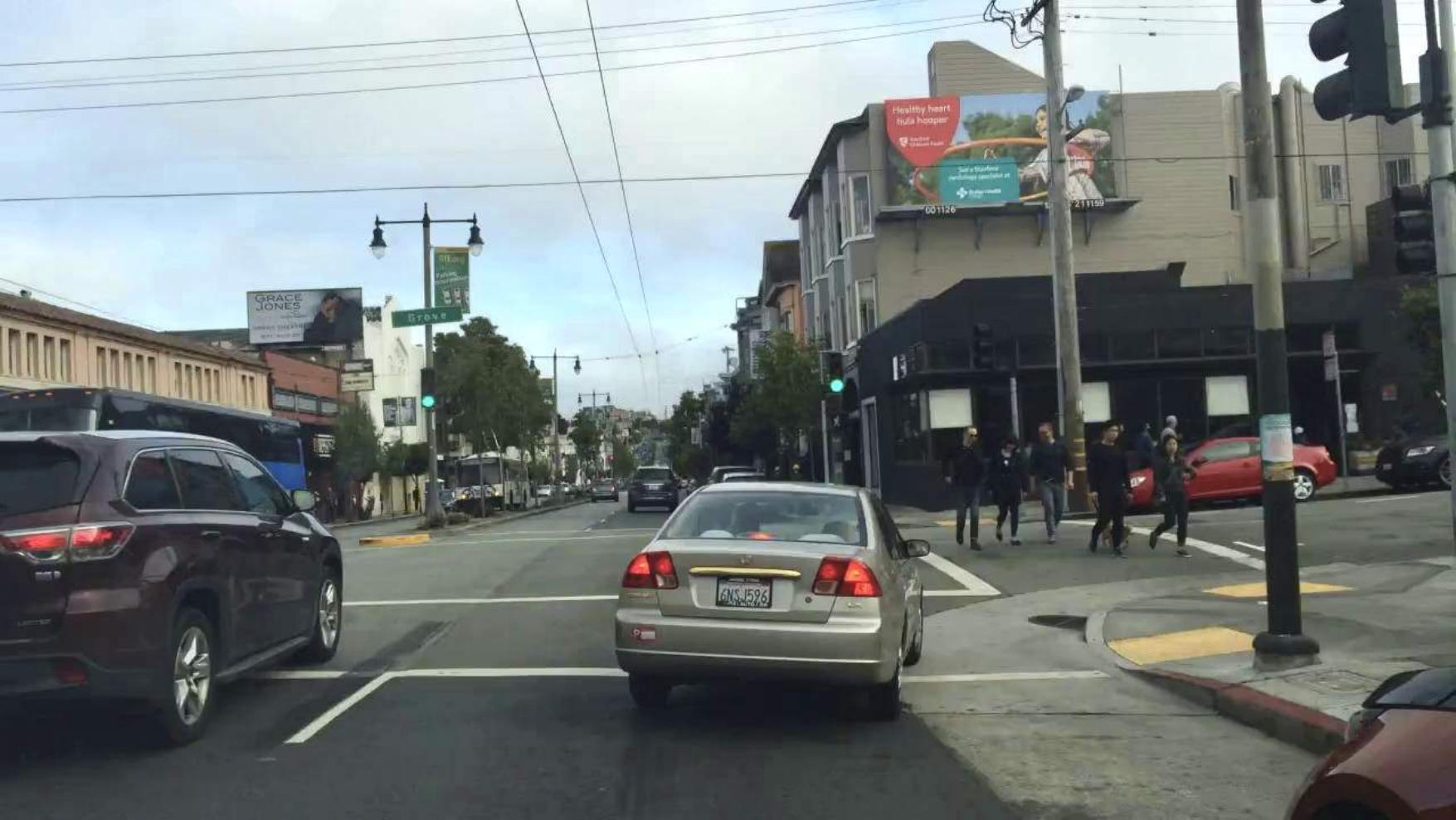}} &    &               \raisebox{-0.5\height}{\includegraphics[width=37.5mm]{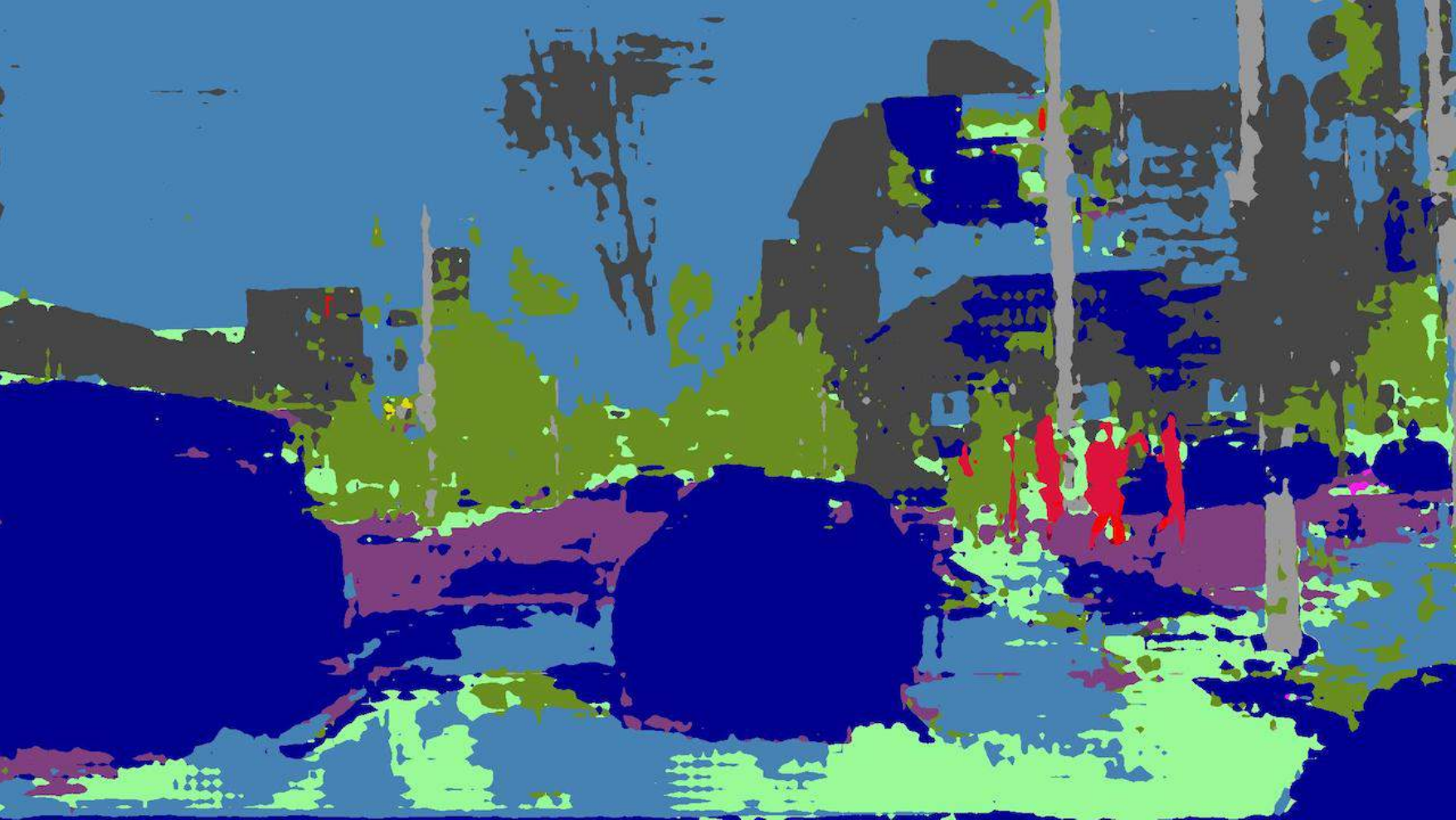}} & 
\raisebox{-0.5\height}{\includegraphics[width=37.5mm]{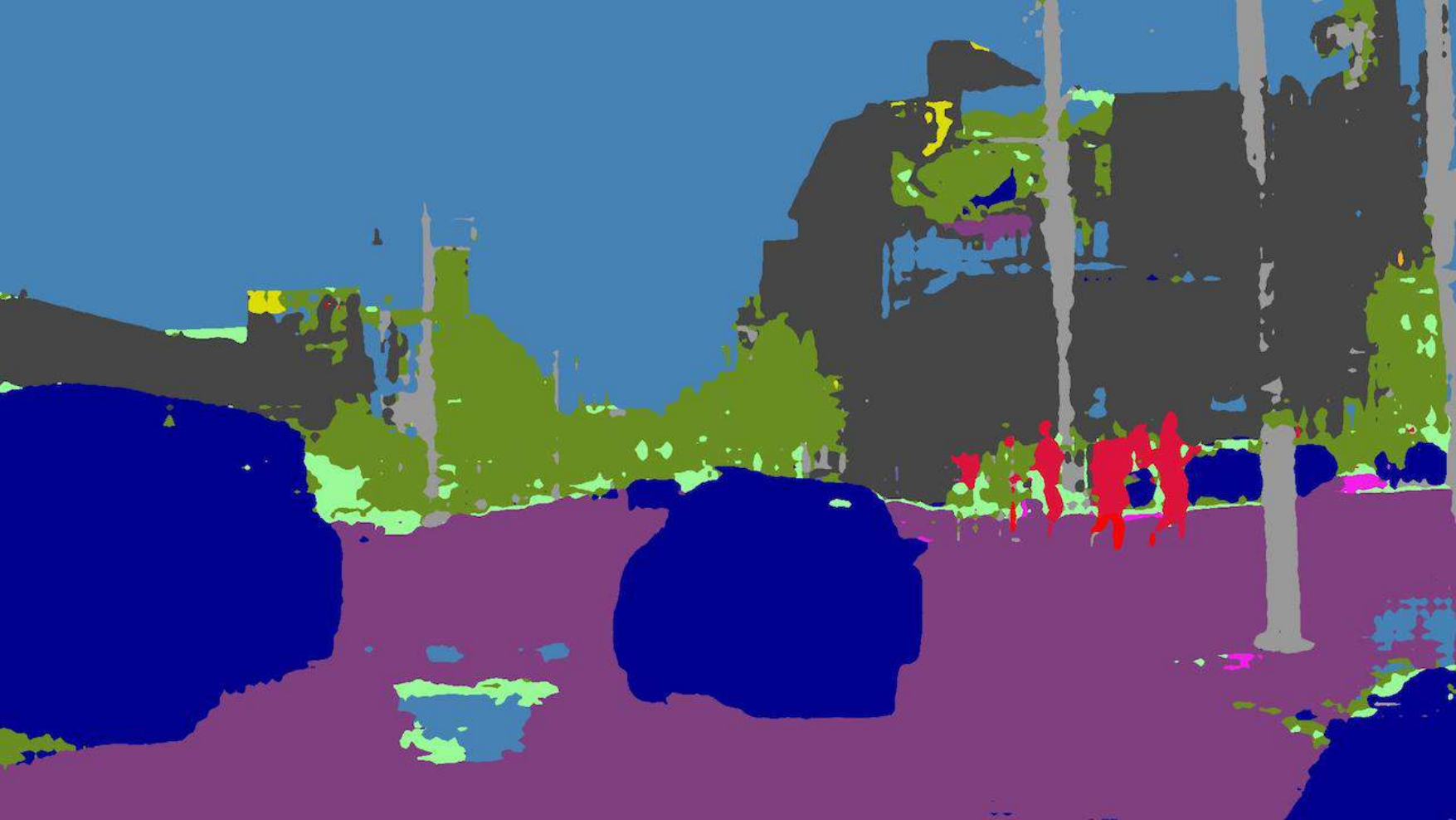}} & 
\raisebox{-0.5\height}{\includegraphics[width=37.5mm]{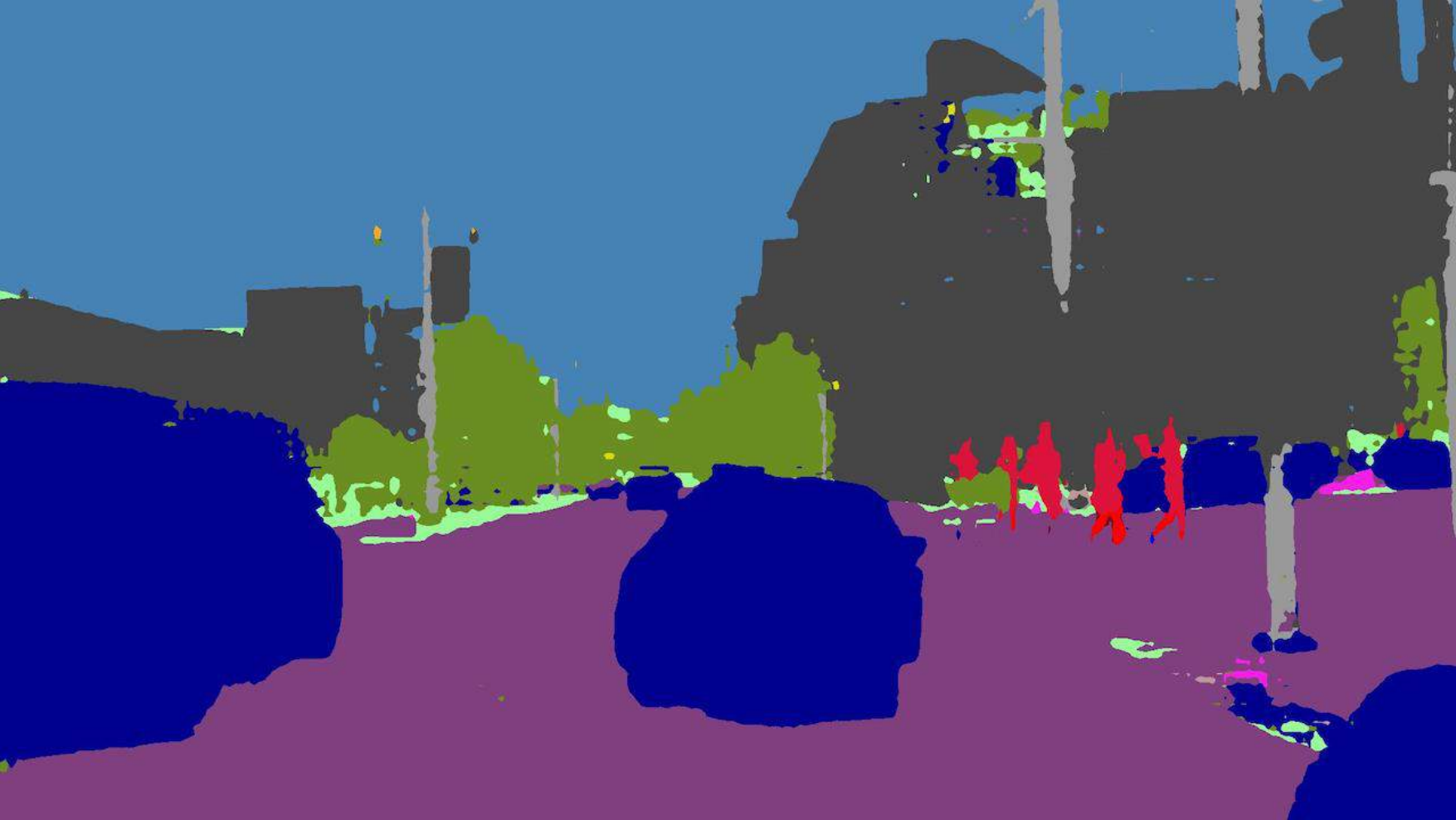}}\\&
i) RGB &   &                
k) DeepLab V2 & 
l) DADA & 
m) ADVENT\\&
\raisebox{-0.5\height}{\includegraphics[width=37.5mm]{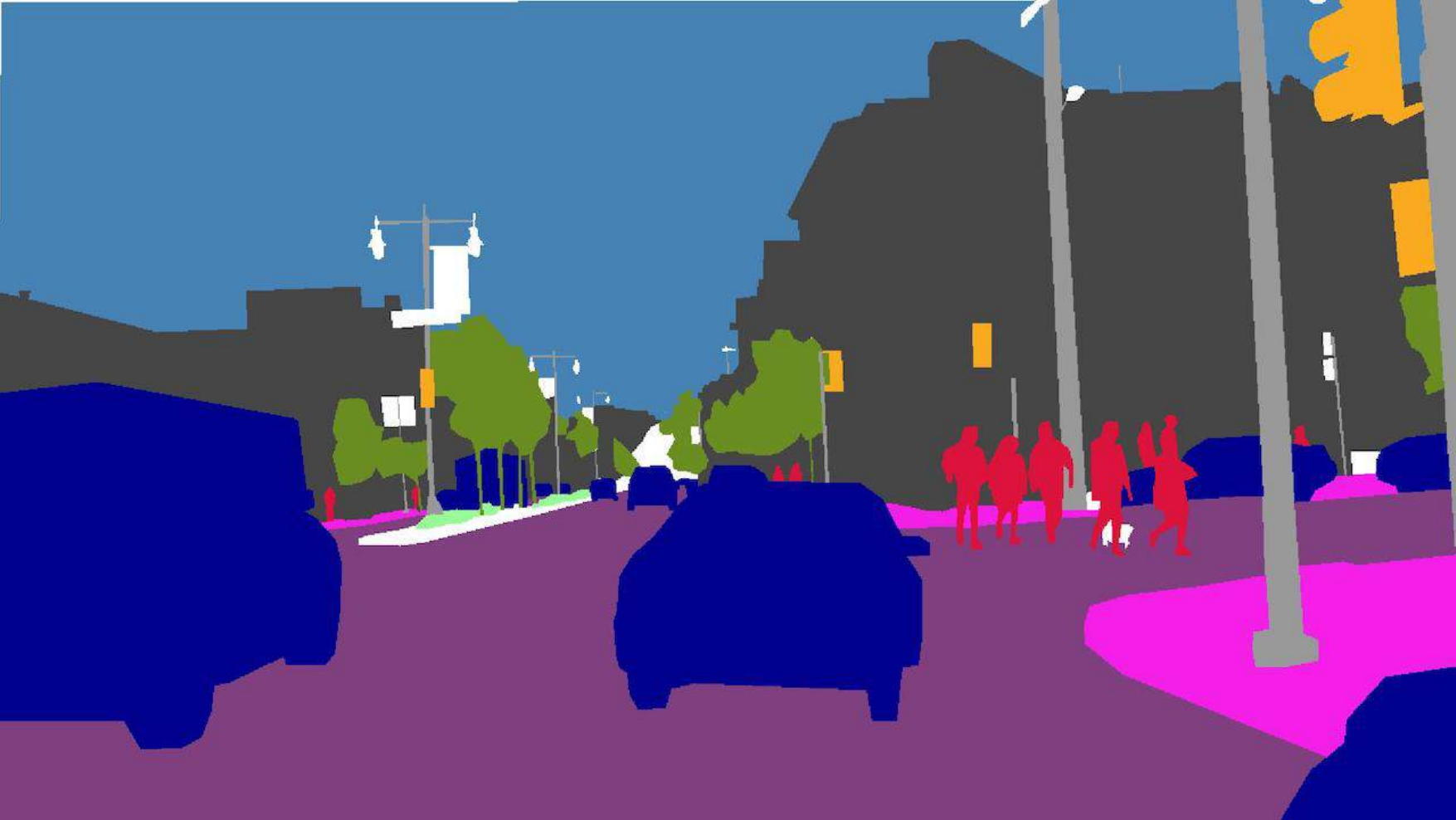}} & &
\raisebox{-0.5\height}{\includegraphics[width=37.5mm]{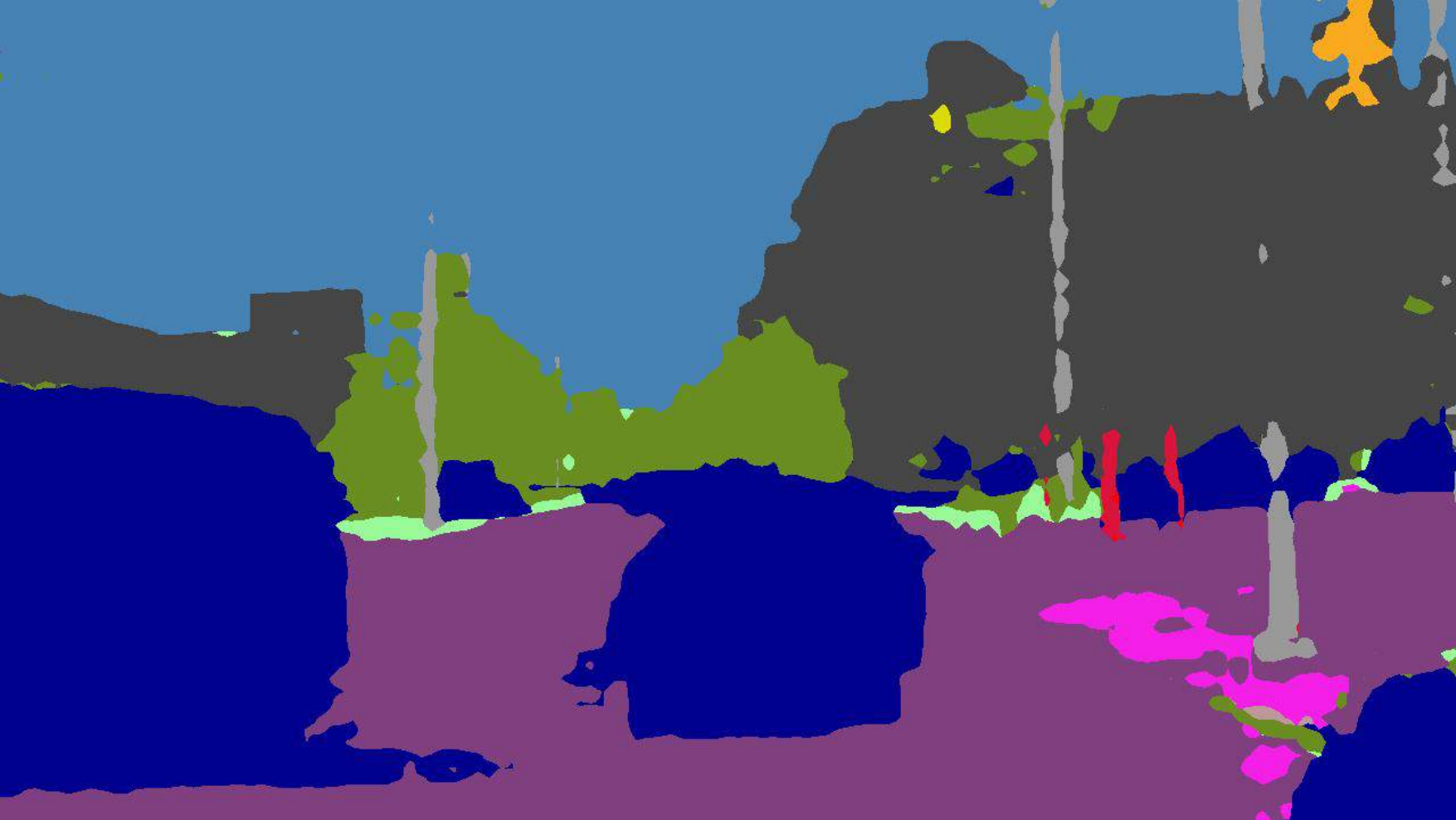}} &
\raisebox{-0.5\height}{\includegraphics[width=37.5mm]{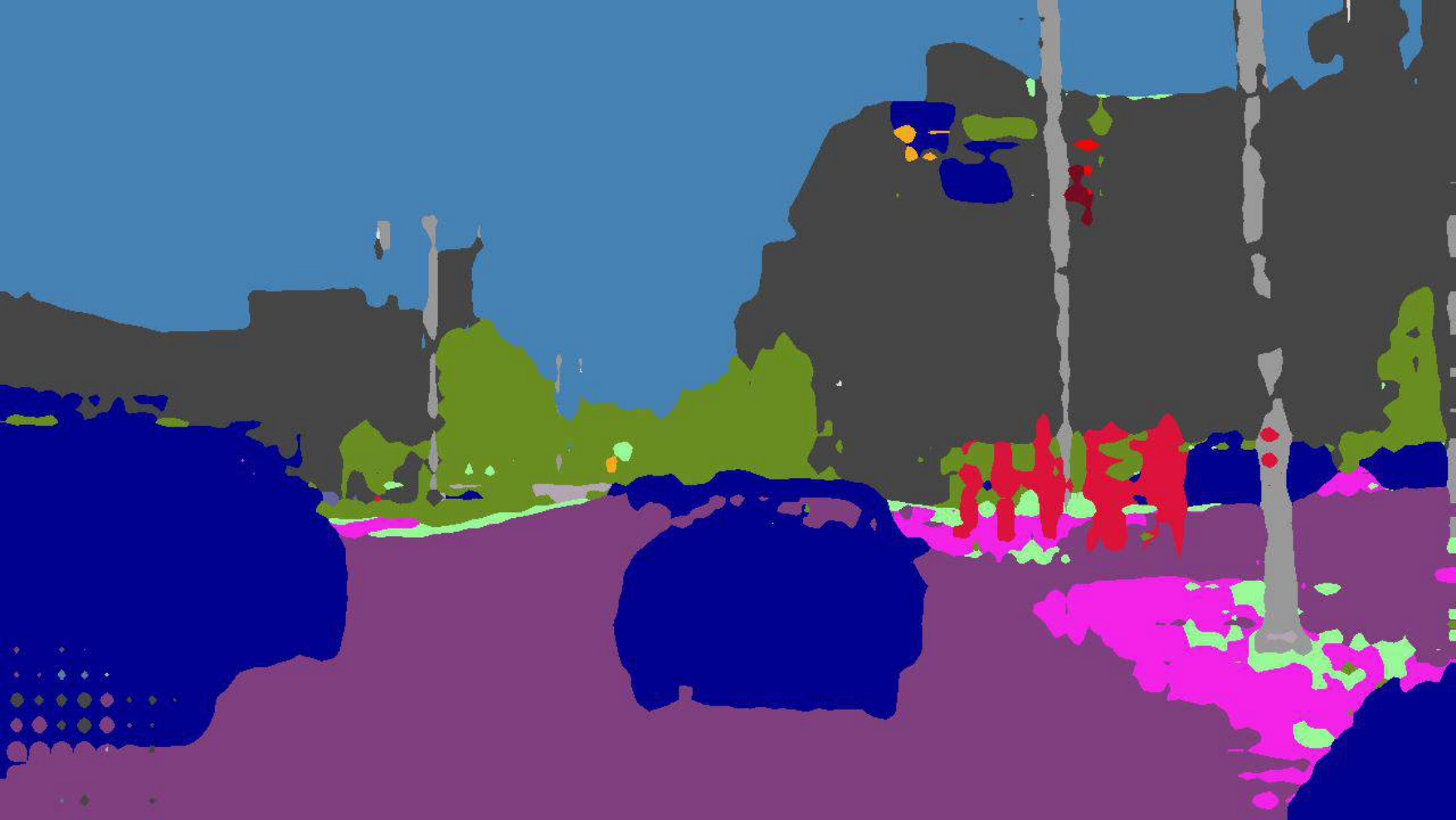}} & 
\raisebox{-0.5\height}{\includegraphics[width=37.5mm]{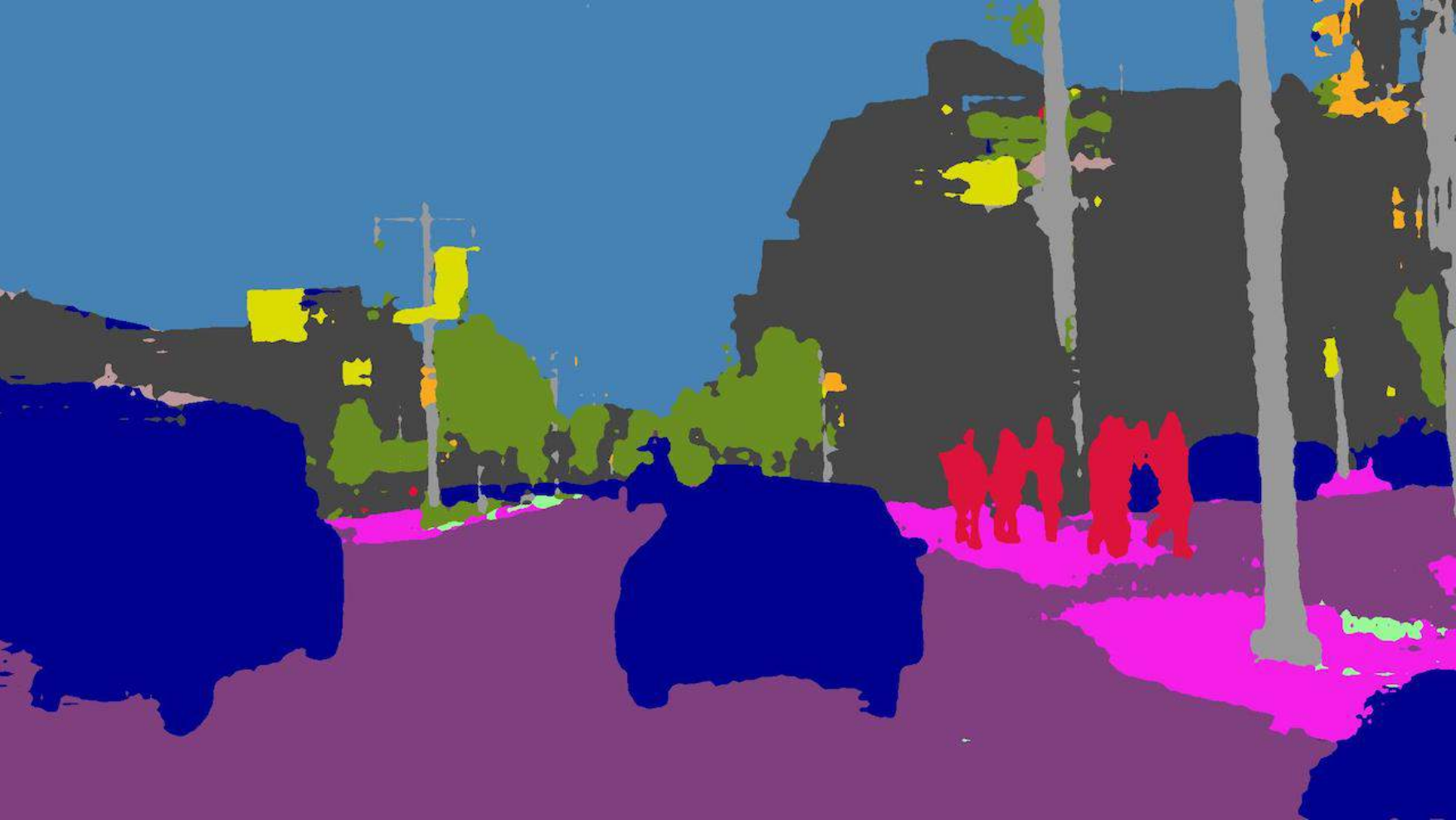}}\\&
j) Ground Truth & &                  
n) CLAN & 
o) DISE & 
p) Baseline
\end{tabular}
\caption{Examples of the results when training on the best and worst case scenarios of IDDA and testing on real datasets. Baseline refers to DeepLab V2 trained on the target (real) domain.}
\vspace{-3mm}
\label{fig:SynVSRealExpImages}
\end{figure*}
With the SemSeg-only architectures we can measure a drop in performance of 30.97\% on average in the best case (not considering the A2D2 experiments since a fair comparison in terms of mIoU cannot be done due to the different evaluation setup). As it can be seen in Fig. \ref{fig:SynVSRealExpImages}c (best case), the network struggles to disambiguate between building, road and sidewalk, though it does an acceptable job at recognizing pedestrians. Among the DA approaches, DISE proves to be the most effective. Nonetheless, the gap with the baseline is still remarkable and the improvements introduced by DA are not enough to guarantee acceptable performance.
Interestingly, it seems that the additional depth information exploited by DADA is helpful only in Mapillary Vistas. 

\begin{table}[htpb]
%\bigskip
\caption{Synthetic vs Real Experiment Results\protect\linebreak *considering only 13 labels}
\begin{adjustbox}{width=1.0\columnwidth}
\begin{tabular}{@{}llcccc@{}}
\toprule
\multirow{2}{*}{Source}            & \multirow{2}{*}{Networks} & \multicolumn{4}{c}{Target}             \\ \cmidrule(l){3-6} 
                             &                           & Cityscapes & BDD100K & Mapillary Vistas & A2D2* \\ \midrule
               Same as target\\(baseline) & DeepLab V2                & 62.89      & 52.71   & 67.63 & 65.43  \\ \midrule\midrule
\multirow{5}{*}{\shortstack{Best case}}  & DeepLab V2                & 32.66      & 24.18   & 36.09  &  32.10         \\ \cmidrule(l){2-6}& DADA                      & 33.13      & 29.58   & 37.29 & 38.57\\ \cmidrule(l){2-6}
& ADVENT                    & 35.32      & 33.18   & 36.97 & 42.56                        \\ \cmidrule(l){2-6} 
                             & CLAN                      & 39.26      & 33.47   & 39.42 & 44.31\\ \cmidrule(l){2-6} 
                             & DISE                      & 42.07      & 40.09   & 41.70   & 46.73                     \\ \midrule
\multirow{5}{*}{Worst case} & DeepLab V2                & 16.81      & 17.48   & 27.09  & 29.80          \\ \cmidrule(l){2-6} & DADA                      & 23.68      & 23.45   & 32.57  & 36.18          \\ \cmidrule(l){2-6} 
                             & ADVENT                    & 23.83      & 27.04   & 30.26  & 38.57          \\ \cmidrule(l){2-6} 
                             & CLAN                      & 25.75      & 30.70    & 30.88   & 42.71         \\ \cmidrule(l){2-6} 
                             & DISE                      & 31.25      & 31.37   & 33.72  & 45.49
                             \\           \bottomrule
\end{tabular}
\end{adjustbox}
\label{table:SynVsRealExperiments}
\end{table}

As expected, in the worst case the domain shift is much more severe, with a maximum drop of 46.08\% when tested on Cityscapes. In this case, the SemSeg-only network fails to even identify the road, confusing it with the ``terrain'' (see Fig. \ref{fig:SynVSRealExpImages}c, worst case). This can be imputable to the relevant textural differences of source and target domains. The impact of DA is visually high, yet numerically we observe how even the best performing architecture does not get close to the baseline. We also note that in both Cityscapes and BDD100K the best performing DA (DISE) almost doubles the performances of the SemSeg-only architecture, but has a much lower increase of performance in the case of Mapillary. This suggests that the higher the performance of the SemSeg-only networks, the lower the impact of DA. 
Overall the gap remains of 28.96\% on average, showing once again how DA techniques still have to work to achieve adequate results.

Looking at the A2D2 results, the SemSeg-only architecture shows a drop in performance close to 30\% both in the Best and Worst cases. The domain shift in the Worst case is a little less severe than Cityscapes and BDD100K, and more similar to Mapillary. This can be imputed to an higher presence of roads out of town, decreasing the difference among A2D2 and the Worst case distribution.

\section{Conclusion}
This paper presents IDDA, a synthetic database explicitly designed for supporting research in domain adaptive semantic segmentation for autonomous driving. With 105 different domains, it is the largest existing dataset supporting this research. As shown in the experiment section, it lends itself well to the benchmark of wide range of domain adaptation study cases, due to the domain gap that exists both among scenarios inside IDDA and with respect to a real dataset.
Furthermore, the constant development of the simulator allows for a further expansion of the currently available scenarios, for instance by adding a night view, new environments or sensor types (such as LIDAR).

\bibliographystyle{IEEEtran} 
\bibliography{egbib}
\end{document}